\documentclass[conference]{IEEEtran}

\IEEEoverridecommandlockouts

\usepackage{times}
\usepackage{caption}
\usepackage[numbers]{natbib}
\usepackage{multicol}
\usepackage[bookmarks=true]{hyperref}

\usepackage{booktabs} 
\usepackage{multirow}
\usepackage{graphicx}
\usepackage{colortbl}
\usepackage{xspace}
\usepackage{mdframed}
\usepackage{subcaption}

\usepackage{amsmath}
\usepackage{subfiles}
\usepackage{dsfont}
\usepackage{tabularx}
\usepackage{soul}
\usepackage{xcolor}
\usepackage{amssymb}
\usepackage{fancyhdr}

\definecolor{lightgray}{gray}{.90}
\definecolor{lightblue}{rgb}{0.9,0.95,1}
\definecolor{lightgreen}{rgb}{0.9,1,0.95}
\definecolor{stateblue}{rgb}{0.8,0.85,1}

\newmdenv[  
  backgroundcolor=lightgray,  
  hidealllines=true,  
  innerleftmargin=8pt,  
  innerrightmargin=8pt,  
  innertopmargin=8pt,  
  innerbottommargin=8pt,
]{graybox}

\newmdenv[  
  backgroundcolor=lightblue,  
  hidealllines=true,  
  innerleftmargin=8pt,  
  innerrightmargin=8pt,  
  innertopmargin=8pt,  
  innerbottommargin=8pt,
]{bluebox}

\newmdenv[  
  backgroundcolor=lightgreen,  
  hidealllines=true,  
  innerleftmargin=8pt,  
  innerrightmargin=8pt,  
  innertopmargin=8pt,  
  innerbottommargin=8pt,
]{greenbox}

\begin{document}

\setstcolor{red} 

\title{\LARGE \bf ApexNav: An Adaptive Exploration Strategy for Zero-Shot Object Navigation with Target-centric Semantic Fusion}

\author{Mingjie Zhang$^{1, 2}$, Yuheng Du$^{1}$, Chengkai Wu$^{1}$, Jinni Zhou$^{1}$, Zhenchao Qi$^{1}$, Jun Ma$^{1}$, and Boyu Zhou$^{2, \dag}$
\thanks{\textbf{$^{\dag}$ Corresponding Author}}
\thanks{$^{1}$ The Hong Kong University of Science and Technology (Guangzhou), Guangzhou, China.}
\thanks{$^{2}$ Southern University of Science and Technology,
Shenzhen, China}
\thanks{{\tt\footnotesize mzhang472@connect.hkust-gz.edu.cn}}
\thanks{{\tt\footnotesize jun.ma@ust.hk}, {\tt\footnotesize zhouby@sustech.edu.cn}}
\thanks{This work was supported by the National Key Research and Development Project of China under Grant 2023YFB4706600.}
}

\maketitle

\begin{abstract}
Navigating unknown environments to find a target object is a significant challenge. While semantic information is crucial for navigation, relying solely on it for decision-making may not always be efficient, especially in environments with weak semantic cues. Additionally, many methods are susceptible to misdetections, especially in environments with visually similar objects. To address these limitations, we propose ApexNav, a zero-shot object navigation framework that is both more efficient and reliable. For efficiency, ApexNav adaptively utilizes semantic information by analyzing its distribution in the environment, guiding exploration through semantic reasoning when cues are strong, and switching to geometry-based exploration when they are weak. 
For reliability, we propose a target-centric semantic fusion method that preserves long-term memory of the target and similar objects, enabling robust object identification even under noisy detections.
We evaluate ApexNav on the HM3Dv1, HM3Dv2, and MP3D datasets, where it outperforms state-of-the-art methods in both SR and SPL metrics. Comprehensive ablation studies further demonstrate the effectiveness of each module. Furthermore, real-world experiments validate the practicality of ApexNav in physical environments.
The code will be released at \href{https://github.com/Robotics-STAR-Lab/ApexNav}{\textcolor{blue}{https://github.com/Robotics-STAR-Lab/ApexNav}}.
\end{abstract}
\IEEEpeerreviewmaketitle

\section{Introduction}
Object goal navigation (ObjectNav) requires an autonomous agent to navigate to a target object specified by its category in an unknown environment \cite{batra2020objectnavrevisitedevaluationembodied}, which is a fundamental challenge in Robotics and AI. With rapid advancements in Large Language Models (LLMs) and Vision-Language Models (VLMs), which encode rich commonsense knowledge, zero-shot ObjectNav (ZSON) approaches have emerged \cite{chen2023traindragontrainingfreeembodied, gadre2023cows, kuang2024openfmnavopensetzeroshotobject, long2024instructnavzeroshotgenericinstruction, yin2024sgnavonline3dscene, yokoyama2024vlfm, yu2023l3mvn, zhang2024trihelper, zhou2023esc}. These methods significantly advance the field by offering superior generalization and sim-to-real transfer capabilities compared to data-driven approaches \cite{chaplot2020object, ramakrishnan2022poni, ye2021efficient}.

Nonetheless, current ZSON methods still face challenges in efficiency and robustness. 
First, most methods \cite{yin2024sgnavonline3dscene, yokoyama2024vlfm, yu2023l3mvn, zhang2024trihelper} rely heavily on semantic reasoning to explore and search. However, semantic guidance is not always available or reliable. For example, at the beginning of task, the agent may face a blank wall or occlusions, limiting semantic cues and hindering effective semantic-based navigation. Moreover, ambiguous or weak semantic cues can mislead the agent. For example, a plant might appear in various rooms like a bedroom, living room, or bathroom, and on different surfaces like a table, windowsill, or floor, making it hard to infer its location without sufficient context. In summary, over-reliance on semantics can reduce navigation efficiency.
Secondly, some approaches \cite{gadre2023cows, yokoyama2024vlfm} rely on single-frame detections to identify targets, while others \cite{kuang2024openfmnavopensetzeroshotobject, yu2023l3mvn} adopt max-confidence fusion to build semantic maps for target identification.
However, relying solely on single-frame results often lacks robustness. The max-confidence fusion approach also struggles with misdetections, as even sporadic high-confidence misdetections can result in incorrect semantic labels that are hard to correct through subsequent observations. These issues are exacerbated in cluttered scenes with occlusions or similar-looking objects. 
Therefore, improving robustness under noisy detection requires multi-frame integration and more reliable fusion strategies.

\begin{figure}[t]
    \begin{center}
    \includegraphics[width=0.99\columnwidth]{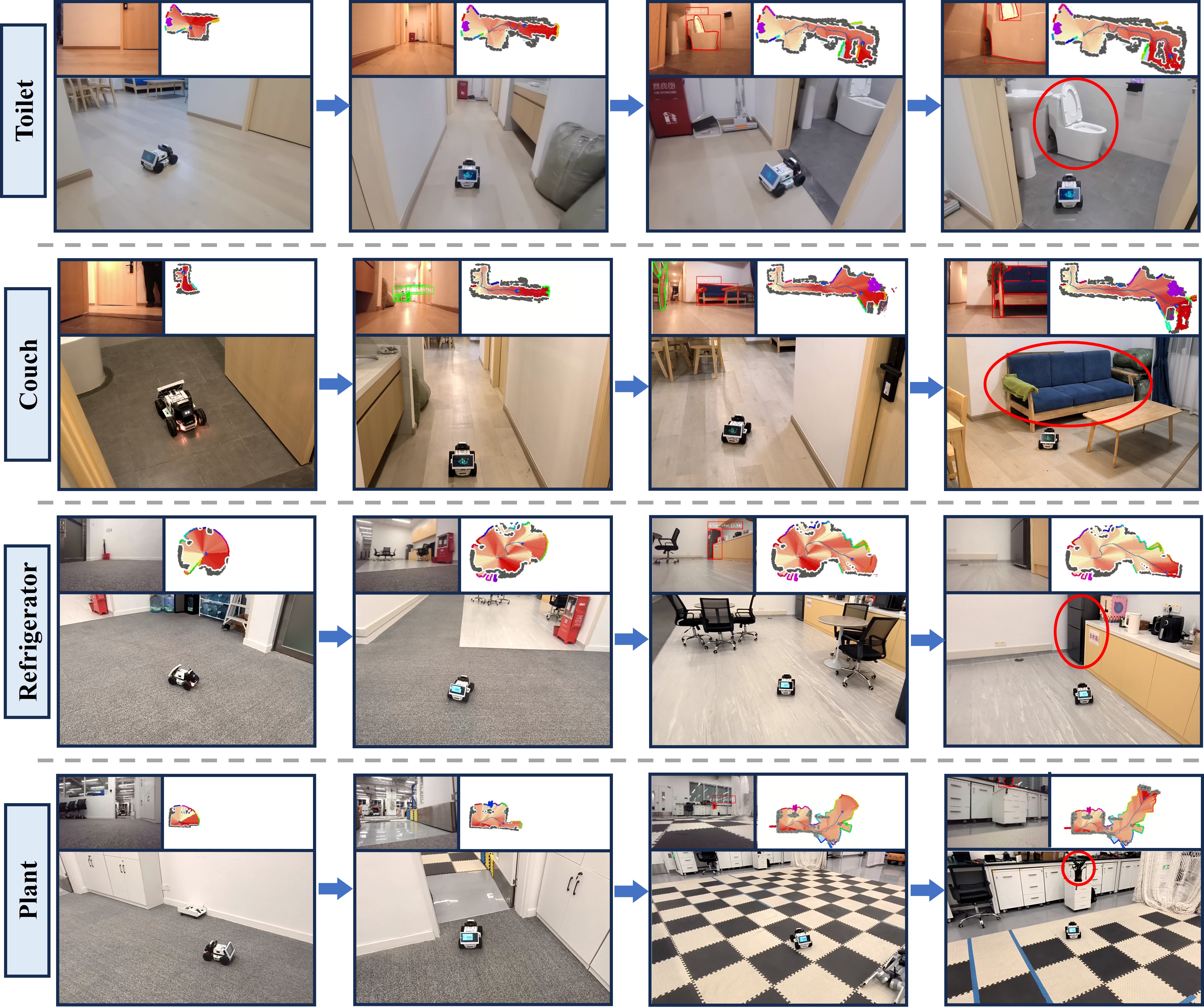}  
    \end{center}
    \vspace{-0.20cm}
    \caption{\label{fig:realworld} \textbf{Real-world Demonstration of ApexNav.} We test ApexNav on various object goals in different environments. The figure shows the navigation process from start to finish, all successfully locating the target objects.}
    \vspace{-0.40cm}
\end{figure}

\begin{figure*}[t]
    \begin{center}
      \includegraphics[width=1.70\columnwidth]{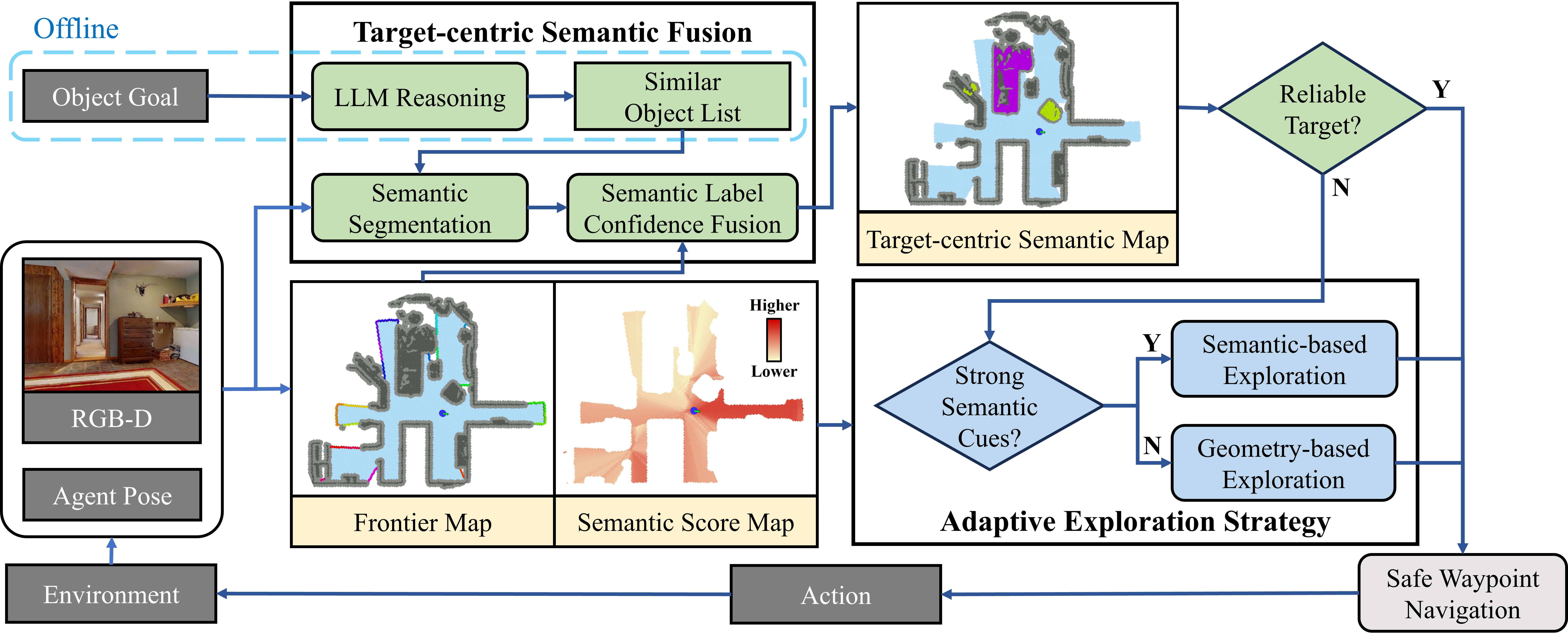}     
    \end{center}
   \vspace{-0.25cm}
   \caption{\label{fig:pipeline} \textbf{System Architecture of ApexNav.} Before the episode, an LLM offline generates a similar object list. The agent builds a frontier map, semantic score map, and a semantic map using target-centric fusion to identify reliable targets. If a reliable target is found, the safe waypoint navigation module directs the agent to it; otherwise, an adaptive exploration strategy selects a frontier waypoint for further exploration.}
   \vspace{-0.50cm}
\end{figure*}

In the paper, we address these limitations by drawing inspiration from human behavior in searching objects. Humans tend to quickly explore their surroundings when semantic cues are weak and concentrate their search in likely target regions when cues are strong. Furthermore, when encountering a potentially target-like object, humans do not make immediate decisions under uncertainty. Instead, they gather additional observations to accumulate evidence and refine their judgment.
Building on these insights, we propose \textbf{ApexNav}, an efficient and reliable framework for ZSON, with two key contributions:

1) \textbf{Adaptive exploration strategy}: ApexNav makes more effective use of semantic information by analyzing the environment's semantic distribution. When strong semantic cues are available, it leverages semantic reasoning to guide the agent toward likely target regions. In contrast, when semantic information is limited, it switches to a geometry-based mode to rapidly explore unknown areas.
To further improve efficiency, semantic-mode frontier selection is formulated as a Traveling Salesman Problem, optimizing the visiting order of high-scoring frontiers and avoiding greedy selection \cite{yin2024sgnavonline3dscene, yokoyama2024vlfm}.

2) \textbf{Target-centric semantic fusion}: ApexNav maintains long-term memory of the target object and visually similar objects. 
By aggregating multi-frame observations with context-aware confidence weighting, it enables reliable target identification even when using noisy and imperfect detectors.
 
We evaluate ApexNav on the HM3Dv1 \cite{ramakrishnan2021habitat}, HM3Dv2 \cite{yadav2023habitat}, and MP3D \cite{chang2017matterport3d} datasets, where it achieves state-of-the-art zero-shot performance.
Notably, it achieves relative improvements in SR and SPL of 5.5\% and 8.6\% on HM3Dv1, and 19.8\% and 16.9\% on HM3Dv2 over previous methods, establishing a new baseline for ZSON. 
We further conduct ablation studies to validate the contribution of each module and analyze failure cases to provide insights for future work.
Extensive real-world experiments also confirm the practicality of ApexNav.

\section{Related Work}

\subsection{Object Goal Navigation}

ObjectNav methods are generally divided into end-to-end and modular approaches. End-to-end methods \cite{ye2021efficient}, based on reinforcement learning require heavy training and have high computational costs. Afterwards, modular learning-based methods \cite{chaplot2020object, ramakrishnan2022poni} reduce training effort but do not fully solve the problem.  
Recent studies introduce LLMs \cite{achiam2023gpt, liu2024deepseek, touvron2023llama} and VLMs \cite{li2023blip, radford2021learning} into ObjectNav, enabling zero-shot methods \cite{chen2023traindragontrainingfreeembodied, gadre2023cows, kuang2024openfmnavopensetzeroshotobject, yin2024sgnavonline3dscene, yokoyama2024vlfm, yu2023l3mvn, zhang2024trihelper, zhou2023esc} that achieve strong performance without additional training.  
Building on this, we further improve efficiency and reliability through an adaptive exploration strategy and a target-centric semantic fusion mechanism.

\subsection{Exploration Strategy for ZSON}

Current ZSON methods mainly use frontier-based exploration. CoW \cite{gadre2023cows} follows a nearest-frontier strategy based only on geometry, ignoring semantic cues in indoor environments. L3MVN \cite{yu2023l3mvn} and TriHelper \cite{zhang2024trihelper} construct a semantic map and use LLMs to identify target-related frontiers, while SG-Nav \cite{yin2024sgnavonline3dscene} leverages 3D scene graphs to prompt LLMs for reasoning. VLFM \cite{yokoyama2024vlfm} applies a VLM to build a value map linking frontiers to the target.
These methods typically score frontiers with LLMs or VLMs and use a greedy policy to select the highest-scoring one. WTRP-Searcher \cite{liu2025handle} instead models ObjectNav as a Weighted Traveling Repairman Problem to overcome greedy policy limitations.
While effective in semantically rich environments, they often perform poorly when semantic cues are weak. SemUtil \cite{chen2023traindragontrainingfreeembodied} notes that early semantic guidance is inefficient and adopts a fixed two-stage strategy that starts with geometry for 50 steps and then switches to semantics. However, this fixed scheme lacks adaptability. Moreover, SemUtil ranks frontiers using a utility function (semantic score divided by path length), the trade-off between semantic relevance and distance may suppress promising frontiers and limit overall performance.
To address these, we propose an adaptive strategy that effectively leverages semantic and geometry information to guide exploration based on the environment's semantic distribution.

\subsection{Target Identification for ZSON}

The success rate of ZSON is directly influenced by accurate target identification, typically using pre-trained detection and segmentation models \cite{liu2025grounding, wang2023yolov7, zhang2023faster} on RGB inputs. As noted in \cite{busch2024one}, false positives from these detectors remain a major limitation.
Some methods \cite{gadre2023cows, yokoyama2024vlfm} rely solely on single-frame detector outputs, making them vulnerable to failure from mispredictions.
Several works \cite{kuang2024openfmnavopensetzeroshotobject, yu2023l3mvn} construct semantic maps via max-confidence fusion, which is prone to high-confidence misdetections that are difficult to correct later. To mitigate this, TriHelper \cite{zhang2024trihelper} and SG-Nav \cite{yin2024sgnavonline3dscene} introduce validation modules: the former leverages VLMs for refinement, while the latter combines multi-frame scores with 3D scene graphs. Although these methods improve reliability, they lack a robust fusion mechanism to effectively manage detector errors in complex environments. 
We propose a target-centric semantic fusion method that aggregates multi-frame observations using context-aware confidence updates, effectively filtering out occasional errors and improving robustness.

\section{Problem Formulation}

We tackle the ObjectNav task \cite{batra2020objectnavrevisitedevaluationembodied}, where an agent must navigate to a specified object (e.g., a "chair") in an unknown environment. The agent uses an egocentric RGB-D camera and an odometry sensor that provides its displacement and orientation relative to the start. The agent is required to efficiently navigate to the target object, minimizing path length and reaching it within a defined success distance.

\section{Methodology}

The overview of ApexNav is shown in Fig.~\ref{fig:pipeline}. The agent constructs a frontier map and a semantic score map (Sec.~\ref{subsec:Environmental Mapping}). Combining the object goal with the similar objects inferred by the LLM, a target-centric semantic fusion method maintains semantic memory and determine the presence of a reliable target (Sec.~\ref{subsec:Target-centric Semantic Fusion}). If a reliable target is found, an object waypoint is generated as the goal. Otherwise, the agent executes an adaptive exploration strategy to select the most suitable frontier waypoint as the goal (Sec.~\ref{subsec:Adaptive Exploration Strategy}). The selected waypoint is then passed to the navigation module, which uses action evaluation to guide movement (Sec.~\ref{subsec:Waypoint Navigation})

\subsection{Environmental Mapping}
\label{subsec:Environmental Mapping}

\subsubsection{\textbf{Frontier Map}} 
\label{subsubsec:frontier_map}

We construct the frontier map on a 2D probabilistic grid, where each cell is labeled as free, occupied, or unknown via raycasting. Depth images are converted to point clouds and denoised using radius outlier removal. A frontier is defined as a free cell next to at least one unknown cell. Following \cite{zhou2021fuel}, we incrementally update frontiers and cluster them using Principal Component Analysis, approximating each cluster by its center to simplify computation.

\subsubsection{\textbf{Semantic Score Map}}
\label{subsubsec:semantic_score_map}

Following \cite{yokoyama2024vlfm}, we build a 2D semantic score map to show the relevance between the environment and the target object, where higher scores mean stronger correlation. Each RGB image and a text prompt are fed into the pre-trained VLM BLIP-2 \cite{li2023blip}, which outputs a cosine similarity score using image-text matching.

We use objects typically associated with specific rooms (e.g., beds in bedrooms) to augment text prompts, which enhances the BLIP-2’s prediction performance.
Prompts like \textit{`Seems there is a \textless target\_object\textgreater or a \textless target\_room\textgreater ahead`} enable BLIP-2 make earlier predictions at longer ranges. Room names are inferred using LLMs. For objects not tied to specific rooms (e.g., plants), prompts only mention the object.
Cosine scores are projected onto observed free grids, weighted by confidence based on viewing angle. Grids near the optical axis get higher confidence, decreasing with angular offset as \( \cos^2\left(\frac{\theta}{\theta_{\text{FoV}}/2} \cdot \frac{\pi}{2}\right) \), where \( \theta \) is the angular offset and \( \theta_{\text{FoV}} \) is the FoV angle.
To integrate semantic information across frames, the updated semantic score \( s^{\text{new}}_{i,j} \) for grid \((i, j)\) is computed as a confidence-weighted average: 
\( s^{\text{new}}_{i,j} = \frac{c^{\text{cur}}_{i,j} s^{\text{cur}}_{i,j} + c^{\text{pre}}_{i,j} s^{\text{pre}}_{i,j}}{c^{\text{cur}}_{i,j} + c^{\text{pre}}_{i,j}} \),  
where \( c^{\text{cur}}_{i,j} \) and \( c^{\text{pre}}_{i,j} \) are the confidence scores from the current and previous frames, respectively. The updated confidence score is computed as  
\( c^{\text{new}}_{i,j} = \frac{(c^{\text{cur}}_{i,j})^2 + (c^{\text{pre}}_{i,j})^2}{c^{\text{cur}}_{i,j} + c^{\text{pre}}_{i,j}} \),  
which emphasizes more confident predictions via squared weighting.

\subsection{Adaptive Exploration Strategy}
\label{subsec:Adaptive Exploration Strategy}

We propose an adaptive exploration strategy that adjusts its behavior based on the availability of semantic cues, ensuring more efficient use of environmental information by leveraging semantic reasoning when cues are strong and relying on geometric exploration when they are weak.

\begin{figure}[t]
 \begin{center}
\includegraphics[width=0.95\columnwidth]{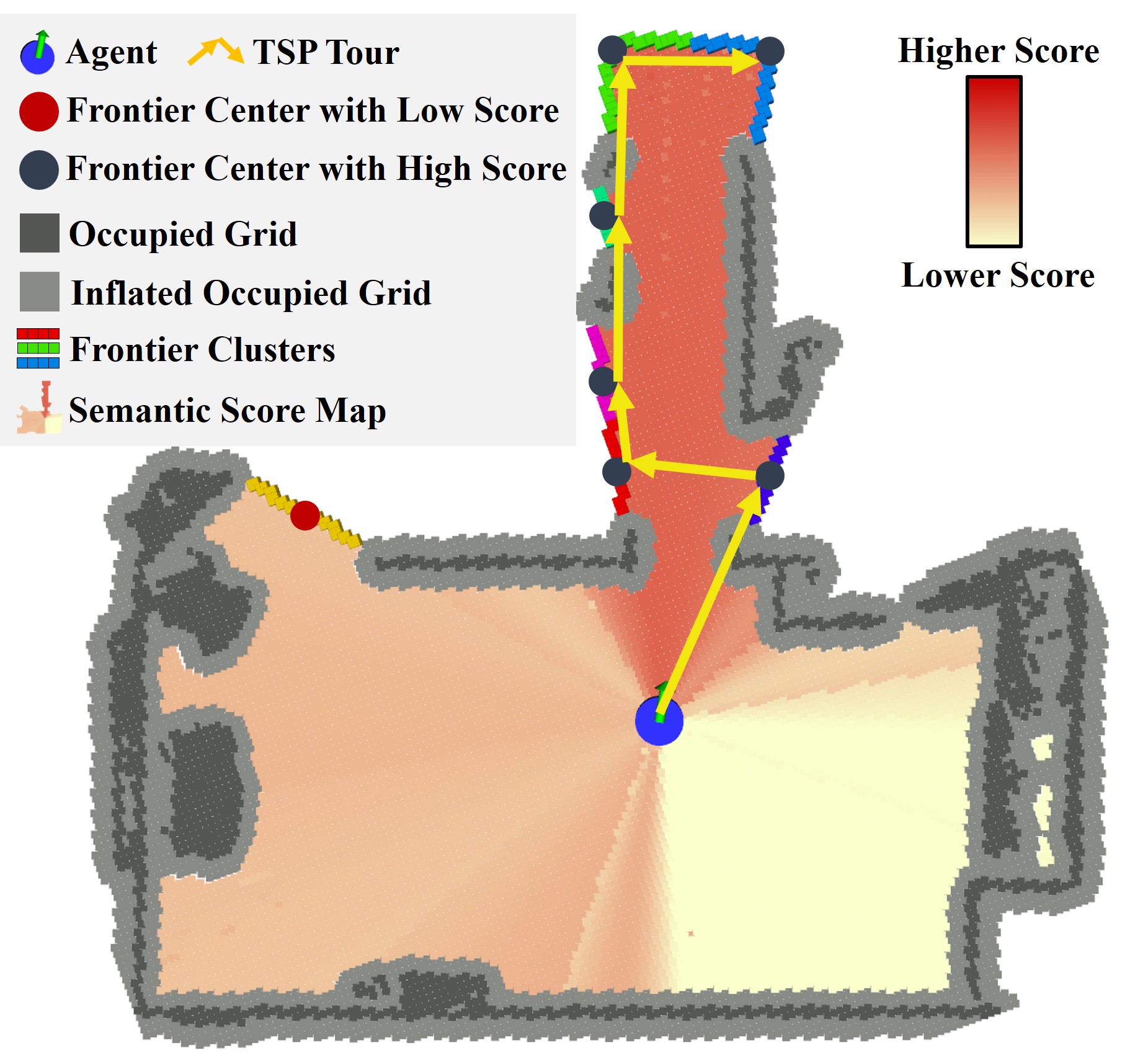}  
 \end{center}
\vspace{-0.25cm}
\caption{\label{fig:semantic_tsp} \textbf{Illustration of Semantic-based Exploration.} Low-score frontier clusters (e.g., the one on the left in the figure) are excluded. A TSP tour is computed over the remaining high-score frontiers, and the agent chooses the first one on the tour as the next navigation goal.}
\vspace{-0.50cm}
\end{figure}

\begin{figure*}[t]
 \begin{center}
\includegraphics[width=1.99\columnwidth]{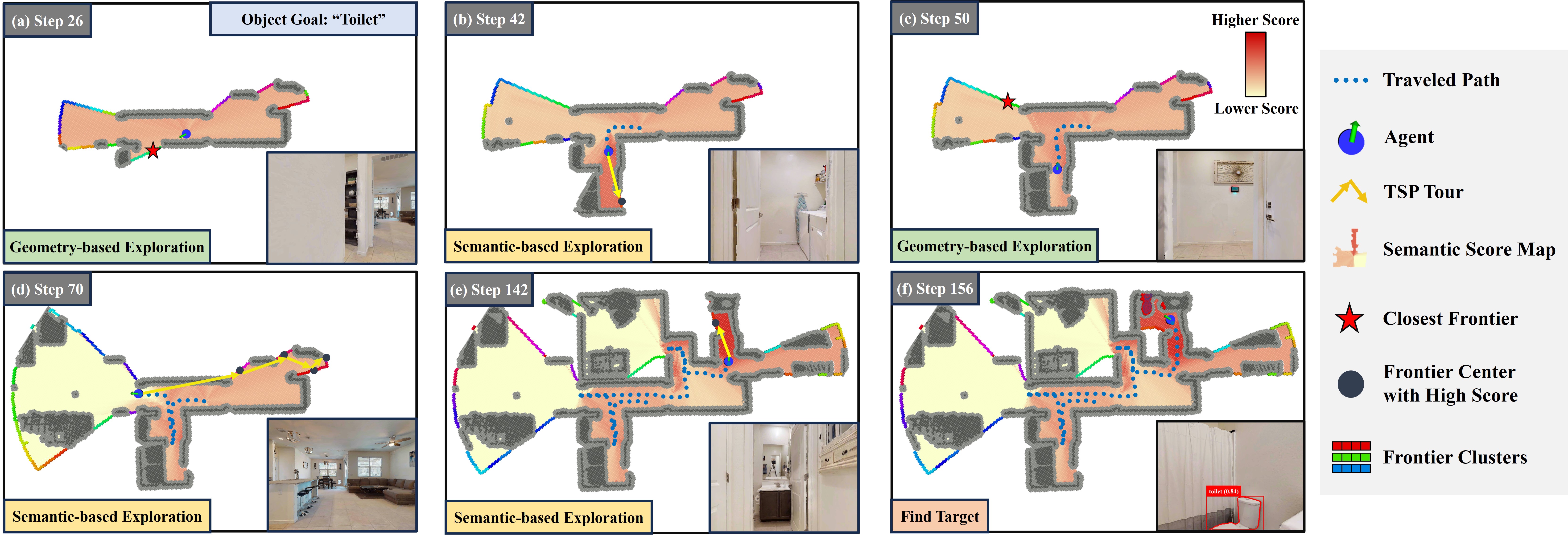}  
 \end{center}
\vspace{-0.15cm}
\caption{\label{fig:adaptive_expl_show} \textbf{An Illustrative Example of Adaptive Exploration.} The agent starts by rotating to initialize the environment. At step 26, it starts searching for the target "toilet." Similar semantic scores on both sides prompt a switch to geometry mode, selecting the nearest frontier. At step 42, it detects washing machines and infers a likely nearby toilet, switching to semantic mode to explore further. When no toilet is found, it returns to geometry mode at step 50 as semantic scores even out. At step 70, it identifies a living room ahead, unlikely to contain a toilet, and switches to semantic mode to explore the high-scoring region on the right. At step 142, it detects a likely bathroom, focuses on the area, and successfully finds the toilet at step 156.}
\vspace{-0.50cm}
\end{figure*}

\subsubsection{\textbf{Mode-switching Criteria}}
\label{subsubsec:Mode-switching Criteria}

To decide the exploration mode, we analyze the semantic score distribution across frontier clusters. Specifically, each frontier is assigned a score based on the semantic score map (Sec.~\ref{subsubsec:semantic_score_map}). We then compute the max-to-mean ratio \( r = s_{\text{max}} / \bar{s} \), where \( s_{\text{max}} \) and \( \bar{s} \) are the maximum and mean scores, respectively. We also calculate the standard deviation \( \sigma = \sqrt{\frac{1}{N_f}\sum_{i=1}^{N_f} (s_i - \bar{s})^2} \), where \( N_f \) is the number of frontier clusters.
A high \(r\) suggests that some frontiers are much more semantically relevant, pointing to promising areas. A high \(\sigma\) also indicates uneven semantic relevance and helps reduce outlier influence, providing a more reliable view of the distribution.

When both \(r\) and \(\sigma\) exceed their respective thresholds (\(r_t\), \(\sigma_t\)), the agent switches to semantic-based exploration, prioritizing high-scoring frontiers to locate the target efficiently. Otherwise, it defaults to geometry-based exploration to expand coverage and gather more environmental information. This adaptive mode-switching mechanism balances rapid geometric exploration with semantic exploration, enhancing the efficiency of the object navigation process.

\subsubsection{\textbf{Semantic-based Exploration}}

When high-score frontiers are present, the system switches to semantic-based exploration. However, always choosing the highest-score frontier can lead to oscillations and inefficient backtracking, especially when scores are similar. To address this, we select a subset of high-score frontiers and solve a Traveling Salesperson Problem (TSP) to plan an optimal visiting sequence. 

We model this problem as an Asymmetric TSP, which can be efficiently solved using the Lin-Kernighan-Helsgaun solver (LKH-Solver) \cite{helsgaun2017extension}, by appropriately designing the cost matrix \(M_{tsp}\). Let \(N_h\) represent the number of high-score frontiers, with their center positions denoted by \(\textbf{p}_i\) (\(i = 1, 2, \dots, N_h\)), and the agent's position as \(p_0\). The cost matrix \(M_{tsp}\) is a $N_{h} + 1$ dimensional square matrix, where the primary \((N_h + 1) \times N_h\) block contains the navigation distances $D(i,j)$ between each frontier and the agent, defined as
\( D(i,j) = \textbf{Len}(\text{Path}(\textbf{p}_{i},\textbf{p}_{j})) \).
Here, \(\text{Path}(\textbf{p}_{i}, \textbf{p}_{j})\) represents the navigation path between \(\textbf{p}_i\) and \(\textbf{p}_j\), computed using the A* algorithm, and \(\textbf{Len}(\text{Path}(\cdot))\) denotes the total length of this path.
The return cost to the agent is set to zero. The cost matrix \(\mathbf{M_{tsp}}(i,j)\) is defined as:
\begin{equation}
     \begin{aligned}
     \mathbf{M_{tsp}}(i,j) =
     \begin{cases}
      0 & \text{if } j = 0, \\
      D(i,j) & \text{if } i \in [0, N_h], j \in (0, N_h] .
     \end{cases}
     \end{aligned}
\end{equation}
Finally, The LKH-Solver computes the optimal path visiting all high-score frontiers, as shown in Fig.~\ref{fig:semantic_tsp}. This approach balances semantic relevance with global path efficiency, reducing redundancy and enhancing exploration consistency.

\subsubsection{\textbf{Geometry-based Exploration}}

When frontier semantic scores are highly uniform, the system switches to geometry-based exploration to quickly acquire new information and expand coverage. As shown in Sec.~\ref{subsec:ablation_exploration}, the nearest-frontier strategy often outperformed TSP-based planning. Although the result was initially unexpected because TSP is typically more efficient in full-coverage tasks \cite{zhou2021fuel}, further analysis revealed that ObjectNav focuses on locating the target rather than performing exhaustive exploration. In this setting, the greedy strategy better aligns with the task objective, leading us to adopt it for geometry-based exploration. Fig.~\ref{fig:adaptive_expl_show} illustrates how our method adaptively switches modes to efficiently locate the target while avoiding irrelevant regions.

\subsection{Target-centric Semantic Fusion}
\label{subsec:Target-centric Semantic Fusion}

\begin{figure}[t]
 \begin{center}
\includegraphics[width=0.95\columnwidth]{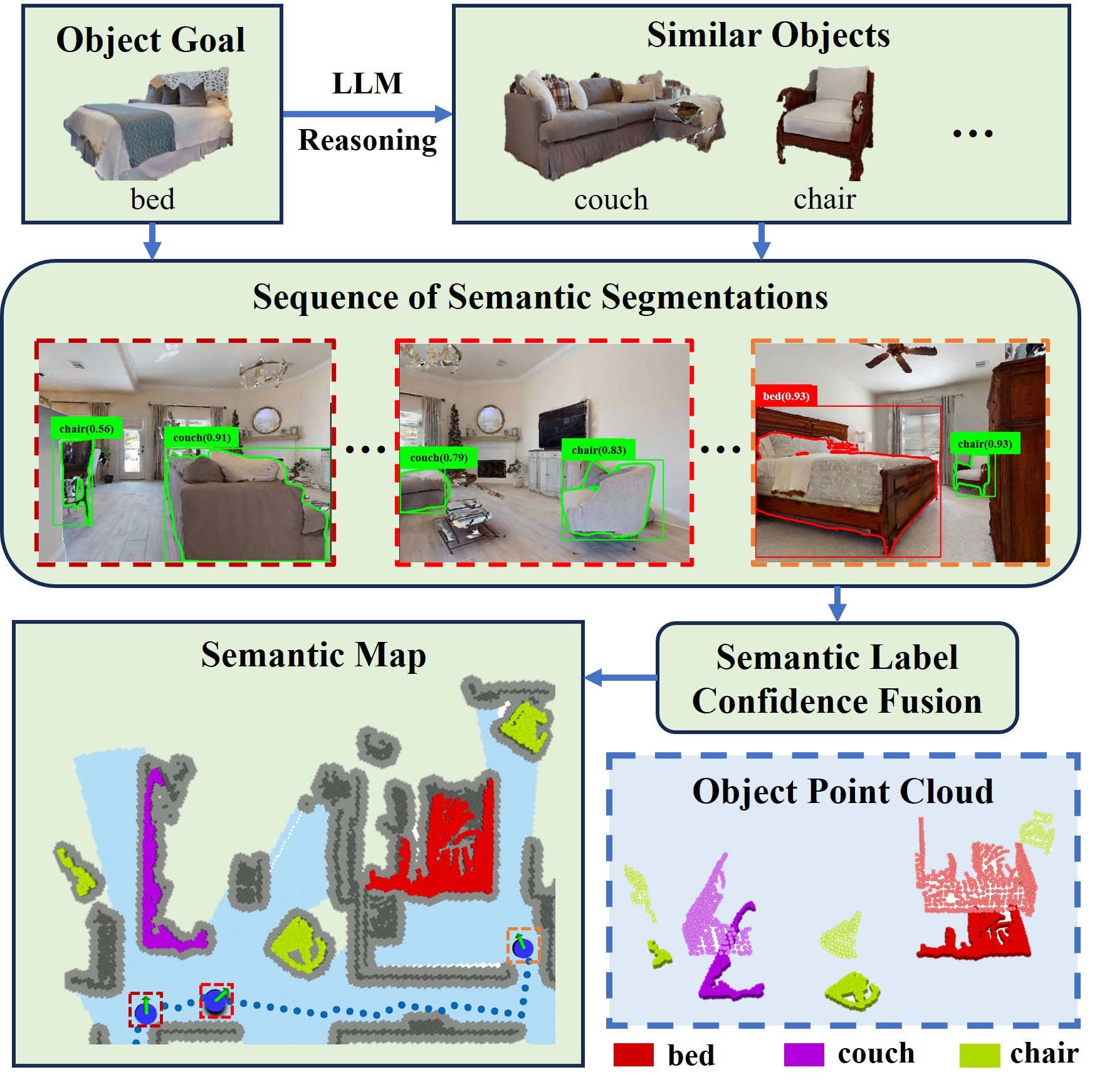}  
 \end{center}
\vspace{-0.25cm}
\caption{\label{fig:fusion_pipeline} \textbf{Pipeline of Target-centric Semantic Fusion.} The agents, enclosed by different-colored dashed boxes, represent semantic segmentations at different time steps. The blue dashed box in the bottom right shows the 3D point cloud information of different objects in the 2D semantic map.}
\vspace{-0.50cm}
\end{figure}

We introduce an object-level semantic fusion method that focuses on the target object and similar ones, improving the reliability of target recognition, as illustrated in Fig.~\ref{fig:fusion_pipeline}.

\subsubsection{\textbf{Object Reasoning with LLMs}}  
\label{subsubsec:Object Reasoning with LLMs}

We leverage the reasoning ability of LLMs with Chain-of-Thought prompting \cite{wei2022chain} to identify objects that are easily misclassified as the target. Considering that detector performance can be affected by object size, as small objects often yield low confidence, the LLM is also prompted to generate an adaptive confidence threshold \(c_{th}\) for the target. An object is treated as a reliable target only if its fused confidence exceeds this threshold. The agent then navigates to the object. This process yields a comprehensive object list \(L_{obj}\) with \(M_{obj}\) entries, including the target and similar objects, guiding subsequent decisions.

\subsubsection{\textbf{Object Detection and Segmentation}}  

The RGB image and the text descriptions of all objects to be detected \(L_{obj}\) are processed by the detector. We use YOLOv7 \cite{wang2023yolov7} for objects within the COCO dataset due to its strong performance, and Grounding-DINO \cite{liu2025grounding} for the rest.  
Detected object instances are segmented by Mobile-SAM \cite{zhang2023faster} using their bounding boxes, and their corresponding point clouds are extracted from the depth map. DBSCAN is applied to remove noise. From a single frame, we obtain \(N_{det}\) detected objects, represented as \( I_{det} = \left\{ \left(pt_{det}^i, c_{det}^i, l_{det}^i\right) \mid i = 1, 2, \dots, N_{det} \right\} \), where \(pt_{det}^i\), \(c_{det}^i\), and \(l_{det}^i\) denote the point cloud, confidence, and label of the \(i\)-th object, respectively.

\subsubsection{\textbf{Semantic Label Confidence Fusion}}

Our fusion method not only incorporates the detector's current detection results but also considers objects that were previously detected but are not identified within the agent's current FoV. To reduce computational overhead and align with our object-level objective, we use objects as the fundamental units for fusion.

\textbf{Structure of Object Clusters:} Each cluster consists of associated 2D grids and maintains object-level information
\(I_o = \left\{ \left(pt_{o}^l, c_{o}^l, n_{o}^l\right) \mid l = 1, 2, \dots, M_{obj} \right\},\) 
where \( M_{obj} \) is the number of object categories (see Sec. \ref{subsubsec:Object Reasoning with LLMs}), and \( l \) is the label index. For each label, the cluster stores the 3D point cloud \( pt_o^l \), confidence \( c_o^l \), and accumulated detection volume \( n_o^l \). This structure allows each cluster to retain multi-label information, offering a comprehensive representation of observed objects.

\textbf{Object Clustering and Fusion:} We fuse single-frame detection results \( I_{det} \) and update object clusters incrementally. Given an arbitrary detection \(I_{det}^i = (pt_{det}^i, c_{det}^i, l_{det}^i)\) from \(I_{det}\), the clustering and fusion process proceeds as follows. The point cloud \(pt_{det}^i\) is projected onto the 2D grid map, and points landing in unoccupied grid cells are filtered out. If no existing object cluster intersects with the projected points, a new object cluster is created with \(M_{obj}\) empty entries, each corresponding to a different label. The entry for label \(l_{det}^i\) is directly filled with the information from \(I_{det}^i\). Otherwise, \(I_{det}^i\) is fused into the intersecting cluster \(O_{last}\), where the pre-fusion information for label \(l_{det}^i\) is \((pt_{last}, c_{last}, n_{last})\), and the post-fusion result becomes \((pt_{new}, c_{new}, n_{new})\).  
Our fusion process weights observations based on detection volume, approximated by the number of points in the down-sampled point cloud, as a larger detected area indicates a more reliable result. Specifically, \( pt_{det}^i \) is down-sampled to a consistent resolution \( r_{down} \), and the resulting number of points in the point cloud, \( v_{det}^i \), represents the detection volume for a single detection result. 
During fusion, \( pt_{det}^i \) is merged with \( pt_{last}^{l_{det}^i} \), and the merged point cloud is down-sampled to generate \( pt_{new}^{l_{det}^i} \). 
The values \( c_{det}^i \) and \( v_{det}^i \) are integrated into \( c_{last}^{l_{det}^i} \) and \( n_{last}^{l_{det}^i} \), updating the accumulated observed points \( n_{new}^{l_{det}^i} \) and the confidence score \( c_{new}^{l_{det}^i} \): 
\begin{equation}
 \begin{split}
  n_{new}^{l_{det}^i} &= n_{last}^{l_{det}^i} + v_{det}^i, \\
  c_{new}^{l_{det}^i} &= \frac{n_{last}^{l_{det}^i}}{n_{new}^{l_{det}^i}} \cdot c_{last}^{l_{det}^i} + \frac{v_{det}}{n_{new}^{l_{det}^i}} \cdot c_{det}^i .
 \end{split}
 \label{eq:fusion}
\end{equation}  

\begin{figure}[t]
 \begin{center}
  \begin{subfigure}[b]{0.99\columnwidth}
    \includegraphics[width=\textwidth]{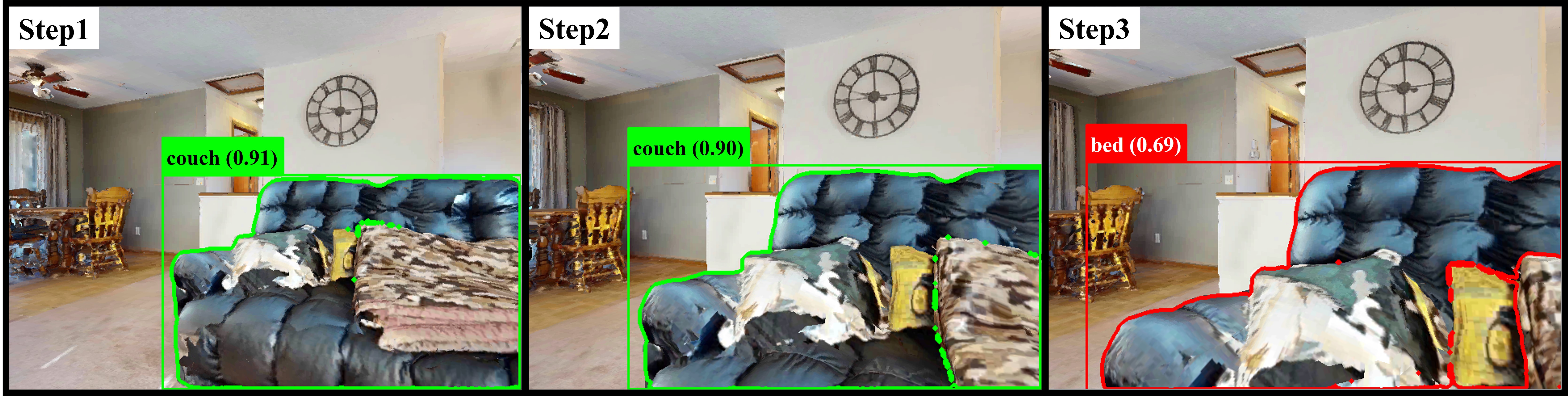}
    \vspace{-0.4cm}
    \caption{Case 1: The target is a bed. At step 3, the detector misclassifies a couch as the bed. However, as the object cluster was already fused twice in steps 1 and 2 with higher-confidence couch data, the misclassification has no effect.}
    \vspace{0.15cm}
\label{fig:fusion_case1}
  \end{subfigure}
  \begin{subfigure}[b]{0.99\columnwidth}
    \includegraphics[width=\textwidth]{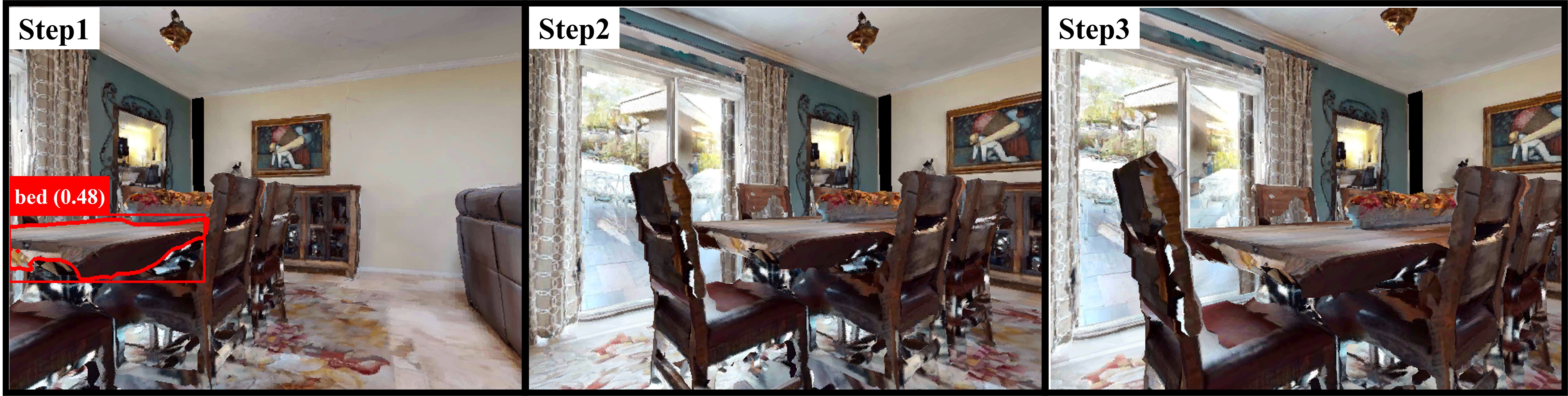}
    \caption{Case 2: The target is a bed. The agent misclassifies a table as a bed in step 1. With its confidence score exceeding the reliable threshold, the agent navigates to the target. However, the object is not detected in steps 2 and 3, the confidence drops below the threshold, so the agent abandons the target.}
\label{fig:fusion_case2}
  \end{subfigure}
 \end{center}
 \vspace{-0.25cm}
 \caption{\label{fig:fusion_cases} \textbf{Two Cases Solvable by Target-centric Semantic Fusion.}}
 \vspace{-0.40cm}
\end{figure}

We also consider object clusters in the current FoV that are missing from the current detection. Such absence may suggest incorrect associations, leading to a confidence penalty. For each label in such clusters, the detection confidence \( c_{det}^{'} \) is set to 0. The fusion weight of this detection is also influenced by the detection volume \( v_{det}^{'} \), determined by the number of points in the point cloud of each label within the object cluster that have a nearest neighbor within the threshold \( r_{down} \) in the point cloud derived from the current observation's depth image.
\( c_{det}^{'} \) and \( v_{det}^{'} \) are fused using Equation~\ref{eq:fusion} for each label within the object cluster, effectively lowering the confidence.
This approach effectively mitigates abnormal confidence scores caused by occasional false positives, as illustrated in Fig.~\ref{fig:fusion_case2}.

\textbf{Determine the Best-matching Label:} After fusion, each object cluster contains information for multiple labels. We select the label with the highest \( c_o \cdot n_o \) as the best match, where \( c_o \) is the confidence score and \( n_o \) is the detection volume, balancing reliability and observation count.

\subsubsection{\textbf{Selection of Reliable Targets}}

Only object clusters whose best-matching label matches the target are considered. Those with confidence above \(c_{th}\) are treated as \textit{reliable targets}, and the agent navigates to them while continuing fusion. Clusters below the threshold are marked as \textit{suspected targets} and are only considered after no frontiers to explore, with the most confident one selected as the goal.

\subsection{Safe Waypoint Navigation}
\label{subsec:Waypoint Navigation}

After initialization, the agent selects either a frontier or an object waypoint as the navigation goal, depending on whether a reliable target is present. Under the Habitat setting, the agent must choose from a set of discrete actions. Many previous methods \cite{chaplot2020object, kuang2024openfmnavopensetzeroshotobject, yin2024sgnavonline3dscene} compute the shortest path and follow it directly; however, these paths often run too close to obstacles, increasing the risk of collisions. To address this, we evaluate actions by jointly considering both efficiency and safety.

\subsubsection{\textbf{Efficiency Cost}}

We use A* to find the shortest path \(P\) and select the first point beyond \(d_{local}\) as the local target \(p_{local}\). Two cost terms guide action selection: the distance cost \(cost_{target}\), which encourages moving toward \(p_{local}\):
\begin{equation}
 cost_{target} = \omega_t \cdot \|p_{new} - p_{local}\|.
\end{equation}  
The proximity change cost \(cost_{prox}\) penalizes actions that increase the distance to \(p_{local}\):  
\begin{equation} 
 \begin{split} 
cost_{prox} =& \omega_p \cdot \left( \|p_{new} - p_{local}\| - \|p_{cur} - p_{local}\| \right). \\  
& \cdot \mathds{1}_{\|p_{new} - p_{local}\| > \|p_{cur} - p_{local}\|} 
 \end{split} 
\end{equation}  
Here, \(\mathds{1}_{\{\cdot\}}\) denotes the indicator function, which returns 1 if the condition in the subscript is met and 0 otherwise.

\subsubsection{\textbf{Safety Cost}}

To evaluate action safety, we build a local ESDF map and compute a safety cost \(cost_{safe}\) by sampling \(K\) points along the action path:
\begin{equation} 
 \begin{split}
 cost_{safe} = \sum_{i=1}^{K} \frac{\omega_s}{d_{obs}(p_{sa_i}) + \epsilon} \cdot \mathds{1}_{d_{obs}(p_{sa_i}) < d_{th}}.
 \end{split}
\end{equation}
Here, \(d_{obs}(p_{sa_i})\) is the distance to the nearest obstacle. Only points within a threshold \(d_{th}\) add to the cost, penalizing unsafe actions. \(\epsilon\) avoids division by zero, and \(\omega_s\) scales the cost.

Finally, the action with the lowest total cost is selected as the agent's next move, ensuring efficient and safe navigation. Notably, in real-world settings where discrete actions are no longer applicable, we employ MINCO \cite{wang2022geometrically} to generate a safe and continuous spatiotemporal trajectory.

\section{Experiments}

\subsection{Simulation Experimental Setup}

\textbf{Datasets:} We evaluate methods in Habitat \cite{puig2023habitat} simulator on three large-scale datasets: \textit{HM3Dv1} (HM3D-Semantics-v0.1 \cite{ramakrishnan2021habitat} from 2022 Habitat Challenge, 2000 episodes, 20 scenes, 6 goal categories), \textit{HM3Dv2} (HM3D-Semantics-v0.2 \cite{yadav2023habitat} from 2023 Habitat Challenge, 1000 episodes, 36 scenes, 6 goal categories), and \textit{MP3D} (Matterport3D \cite{chang2017matterport3d} from 2021 Habitat Challenge, 2195 episodes, 11 scenes, 21 goal categories).

\textbf{Evaluation Metrics:} We report Success Rate (SR) and Success weighted by inverse Path Length (SPL) for all methods, where SR measures episode success and SPL assesses efficiency relative to the optimal path (SPL is 0 for failures). 
For ablation studies, we also report SoftSPL, which accounts for goal-directed progress in failed episodes. 
Higher SR, SPL, and SoftSPL indicate superior performance.

\begin{table}[t]
    \centering
    \setlength\tabcolsep{2.0pt}
    \resizebox{0.49\textwidth}{!}{
    \begin{tabular}{ccccccccc}
    \toprule
    \multirow{2}*{\textbf{Method}} & \multirow{2}*{\textbf{Zero-shot}} &
    \multicolumn{2}{c}{\textbf{HM3Dv1}} & \multicolumn{2}{c}{\textbf{HM3Dv2}} & \multicolumn{2}{c}{\textbf{MP3D}} \\
    \cmidrule(lr){3-4} \cmidrule(lr){5-6} \cmidrule(lr){7-8}
    & & SR$\uparrow$ & SPL$\uparrow$ & SR$\uparrow$ & SPL$\uparrow$ & SR$\uparrow$ & SPL$\uparrow$ \\
    \midrule
    SemExp\cite{chaplot2020object} & $\times$ & -- & -- & -- & -- & 36.0 & 14.4 \\
    PONI\cite{ramakrishnan2022poni} & $\times$ & -- & -- & -- & -- & 31.8 & 12.1 \\
    ZSON\cite{majumdar2022zson} & $\times$ & 25.5 & 12.6 & -- & -- & 15.3 & 4.8 \\
    \midrule
    CoW\cite{gadre2023cows} & $\sqrt{}$ & -- & -- & -- & -- & 7.4 & 3.7 \\
    ESC\cite{zhou2023esc} & $\sqrt{}$ & 39.2 & 22.3 & -- & -- & 28.7 & 14.2 \\
    L3MVN\cite{yu2023l3mvn} & $\sqrt{}$ & 50.4 & 23.1 & 36.3 & 15.7 & 34.9 & 14.5 \\
    OpenFMNav\cite{kuang2024openfmnavopensetzeroshotobject} & $\sqrt{}$ & 54.9 & 24.4 & -- & -- & 37.2 & 15.7 \\
    InstructNav\cite{long2024instructnavzeroshotgenericinstruction} & $\sqrt{}$ & -- & -- & 58.0 & 20.9 & -- & -- \\
    VLFM\cite{yokoyama2024vlfm} & \ \ $\sqrt{}\mkern-9mu{\smallsetminus}$
 & 52.5 & \cellcolor{lightblue}30.4 & \cellcolor{lightblue}63.6 & \cellcolor{lightblue}32.5 & 36.4 & \cellcolor{lightblue}17.5\\
    VLFM*\cite{yokoyama2024vlfm} & $\sqrt{}$ & {50.9} & {23.6} & {56.9} & {27.5} & {32.5} & {15.9} \\
    TriHelper\cite{zhang2024trihelper} & $\sqrt{}$ & \cellcolor{lightblue}56.5 & 25.3 & -- & -- & -- & -- \\
    SG-Nav\cite{yin2024sgnavonline3dscene} & $\sqrt{}$ & 54.0 & 24.9 & 49.6 & 25.5 & \cellcolor{stateblue}\textbf{40.2} & 16.0 \\
    \rowcolor{lightgray} \textbf{ApexNav(Ours)} & $\sqrt{}$ & \cellcolor{stateblue}\textbf{59.6} & \cellcolor{stateblue}\textbf{33.0} & \cellcolor{stateblue}\textbf{76.2} & \cellcolor{stateblue}\textbf{38.0} & \cellcolor{lightblue}39.2 & \cellcolor{stateblue}\textbf{17.8} \\
    \bottomrule
     \end{tabular}
 }
 \caption{\textbf{Benchmark Comparisons.} Data with a \colorbox{stateblue}{stateblue background} shows the best result, while \colorbox{lightblue}{lightblue background} indicates the second-best.}
 \label{table:comparison}
 \vspace{-0.6cm}
\end{table}

\begin{figure*}[t]
 \begin{center}
\includegraphics[width=1.80\columnwidth]{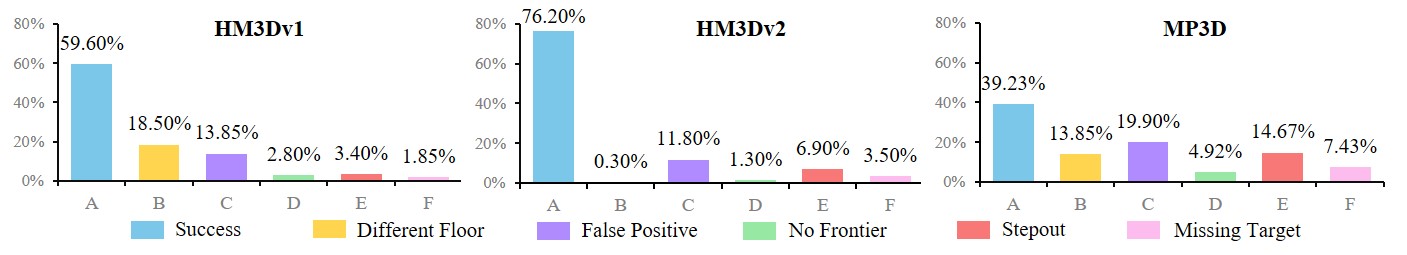}  
 \end{center}
\vspace{-0.25cm}
\caption{Failure Cause Statistics of ApexNav on the Three Datasets.}
\label{fig:Failure Cause Analysis}
\vspace{-0.4cm}
\end{figure*}

\textbf{Implementation Details:} 
In the Habitat setting, the agent selects actions from a discrete set $\mathcal{A}$:  
\textsc{move\_forward} ($0.25\,\text{m}$),  
\textsc{turn\_left} ($30^\circ$),  
\textsc{turn\_right} ($30^\circ$),  
\textsc{look\_up} ($30^\circ$),  
\textsc{look\_down} ($30^\circ$), and  
\textsc{stop}.  
Each episode is limited to 500 steps.
The agent operates within a perception range of \( [0.0\,\text{m}, 5.0\,\text{m}] \), with a success distance of \( 0.2\,\text{m} \). The onboard camera is mounted at a height of \( 0.88\,\text{m} \). For semantic segmentation, we use YOLOv7 \cite{wang2023yolov7}, GroundingDINO \cite{liu2025grounding}, and Mobile-SAM \cite{zhang2023faster}. The semantic score map is generated by BLIP-2 \cite{li2023blip}. DeepSeek-V3 \cite{liu2024deepseek} is used to generate the similar object list, rooms where objects are likely to appear, and detection confidence thresholds.
The parameters are set as follows: \( r_t = 1.10 \), \( \sigma_t = 0.015 \) (Sec.~\ref{subsubsec:Mode-switching Criteria}), and \( d_{\text{local}} = 0.75 \), \( \omega_t = 200 \), \( \omega_p = 2000 \), \( \omega_s = 1.0 \), \( \epsilon = 10^{-2} \), \( K = 10 \), \( d_{\text{th}} = 0.15 \) (Sec.~\ref{subsec:Waypoint Navigation}).
All simulation experiments are run on a system with an NVIDIA GeForce RTX 4090 GPU and an Intel i9-13900K ×32 CPU.

\textbf{Baselines:} SemExp \cite{chaplot2020object}, PONI \cite{ramakrishnan2022poni}, and ZSON \cite{majumdar2022zson} rely on task-specific training, limiting their generalization in zero-shot settings.
CoW \cite{gadre2023cows} uses a nearest-frontier exploration strategy without semantic reasoning.
ESC \cite{zhou2023esc}, L3MVN \cite{yu2023l3mvn}, TriHelper \cite{zhang2024trihelper}, and OpenFMNav \cite{kuang2024openfmnavopensetzeroshotobject} improve exploration by building semantic maps and using LLMs to select frontiers.
VLFM \cite{yokoyama2024vlfm} and InstructNav \cite{long2024instructnavzeroshotgenericinstruction} use LLMs or VLMs to create value maps that guide agents. {VLFM* replaces the pre-trained PointNav module with a shortest-path planner for fair comparison.}
SG-Nav \cite{yin2024sgnavonline3dscene} further enhances semantic reasoning by prompting LLMs with 3D scene graphs to select frontiers.

\subsection{Comparison with State-of-the-art}

Table~\ref{table:comparison} compares ApexNav with SOTA methods on HM3Dv1, HM3Dv2, and MP3D datasets. Blank cells indicate that the method was not evaluated on that dataset. As only InstructNav reported HM3Dv2 results, we re-evaluated several open-source methods under our settings for fairness.
ApexNav demonstrates superior performance over all the SOTA methods. On HM3Dv1, ApexNav achieves a 5.5\% improvement in SR compared to the suboptimal TriHelper and an 8.6\% enhancement in SPL relative to VLFM. On HM3Dv2, ApexNav exhibits a 19.8\% increase in SR and a 16.9\% improvement in SPL compared to the previously best-performing method, VLFM. On MP3D, while the SR of ApexNav is marginally lower than that of SG-Nav, it achieves an 11.3\% higher SPL compared to SG-Nav and a 1.7\% improvement compared to suboptimal VLFM. 
The great enhancement in SR validates the efficacy of the target-centric semantic fusion mechanism introduced by ApexNav in rectifying error detection, while the notable improvement in SPL is predominantly attributable to the adaptive exploration strategy's superior efficiency and robustness in exploring different environments. 

\begin{table}[t]
    \centering
    \setlength\tabcolsep{6.5pt}
    \resizebox{0.90\columnwidth}{!}{
    \begin{tabular}{cccccc}
         \toprule
         \multicolumn{2}{c}{\textbf{Exploration Strategy}} & \multirow{2}*{SR$\uparrow$} & \multirow{2}*{SPL$\uparrow$} & \multirow{2}*{SoftSPL$\uparrow$} \\
         \cmidrule(lr){1-2}
         Semantic & Geometry & & & \\
         \midrule
         Greedy & -- & 72.90 & 36.07 & 37.85 \\
         TSP & -- & 74.50 & 37.03 & 39.02 \\
         -- & Greedy & 73.80 & 36.42 & 38.40 \\
         -- & TSP & 72.90 & 34.25 & 36.87 \\
         \midrule
         TSP & TSP & 74.20 & 36.79 & 38.88 \\
         Greedy & Greedy & 74.80 & 37.67 & 39.65 \\
         Greedy & TSP & 75.00 & 36.61 & 38.61 \\
         \rowcolor{lightgray}TSP & Greedy & \textbf{76.20} & \textbf{38.03} & \textbf{40.44} \\
         \bottomrule
         \multicolumn{2}{c}{{Utility-based}} & {74.40} & {36.47} & {38.94} \\
         \bottomrule
    \end{tabular}
    }
    \caption{Ablation Study on Exploration Strategy.}
    \label{table:ablation_strategy}
    \vspace{-0.5cm}
\end{table}

\subsection{Failure Cause Analysis}
\label{subsec:failure cause analysis}

Task outcomes are categorized as follows:

\begin{itemize}

    \item \textbf{Success (A):} The agent stops within $0.2m$ of the goal and the simulator confirms success.
    
    \item \textbf{Different Floor (B):} All target objects are on a different floor from the agent’s start position.
    
    \item \textbf{False Positive (C):} The agent stops near a detected object, but the simulator returns failure, meaning the wrong object was found.
    
    \item \textbf{No Frontier (D):} The agent runs out of frontiers to explore without passing the target.
    
    \item \textbf{Stepout (E):} The agent exceeds the 500-step limit without passing the target.
    
    \item \textbf{Missing Target (F):} Passes the target but fails to recognize it, either due to no frontiers left or step limit reached.
    
\end{itemize}

As shown in Fig.~\ref{fig:Failure Cause Analysis}, "False Positive" cases account for over 11\% across all datasets, mainly due to persistent detection errors and incorrect fusion. Another major cause is annotation issues, where the agent finds the target but the task still fails because the object is not labeled.

Except for the HM3Dv2 dataset, which mainly consists of single-floor tasks, the others show many "Different Floor" failures (\(>13\%\)). {This is because ApexNav builds only a single-layer 2D map, so such cross-floor tasks are treated as failures by default.}
This limitation also leads to many "No Frontier" failures. When the agent starts at a staircase between two floors, the vertical distance to the target is not large enough to be recognized as on a different floor. However, since the agent builds a 2D map centered at its starting position, it may find no frontiers to explore after initialization, resulting in task failure. {This limitation is mainly an engineering constraint, and could be addressed by incorporating 3D maps in future work.}

\begin{table}[t]
    \centering
    \setlength\tabcolsep{3.5pt}
    \resizebox{0.95\columnwidth}{!}{
    \begin{tabular}{ccccc}
         \toprule
         \textbf{Serial} & \textbf{Exploration Metrics} & SR$\uparrow$ & SPL$\uparrow$ & SoftSPL$\uparrow$ \\
         \midrule
         i & \( \sigma_t = 0.000, r_t=1.00 \) & 74.50 & 37.03 & 39.02 \\
         ii & \( \sigma_t = 0.010, r_t=1.05 \) & 73.70 & 37.80 & 39.97 \\
         \rowcolor{lightgray} iii & \(\sigma_t = 0.015, r_t = 1.10\) & \textbf{76.20} & \textbf{38.03} & \textbf{40.44} \\
         iv & \( \sigma_t = 0.020, r_t=1.15 \) & 75.40 & 37.76 & 40.17 \\
         v & \( \sigma_t = 0.025, r_t=1.20 \) & 74.90 & 37.42 & 40.05 \\
         vi & \( \sigma_t = \infty, r_t=\infty \) & 73.80 & 36.42 & 38.40 \\
         \bottomrule
    \end{tabular}
    }
    \caption{Ablation Study on Exploration Metrics.}
    \label{table:ablation_metrics}
    \vspace{-0.20cm}
\end{table}

\begin{figure}[t]
    \begin{center}
        \includegraphics[width=0.85\linewidth]{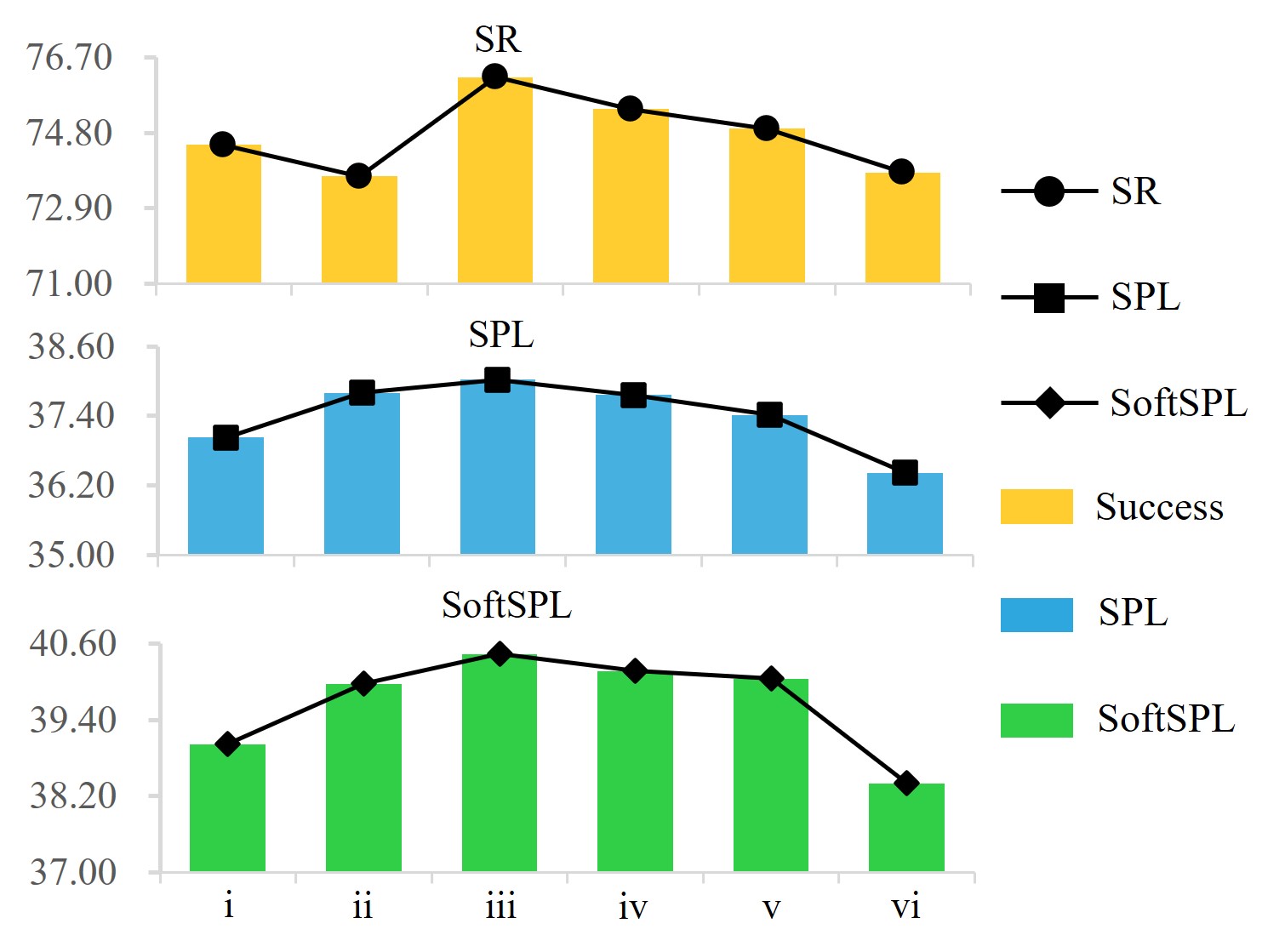}
    \end{center}
    \vspace{-0.2cm}
    \caption{Ablation Study on Exploration Metrics. }
    \label{fig:ablation_metrics}
    \vspace{-0.6cm}
\end{figure}

“Stepout” failures are more common in HM3Dv2 and MP3D due to their larger scene scale, which make it hard for the agent to find the target within 500 steps. Raising the step limit could help. Additionally, some scenes in MP3D have black holes that cause missing depth data, leading to map mismatches and the agent getting stuck. We recommend focusing on datasets with more complete scene structures, like HM3Dv2, for future research.
“Missing Target” cases are rare and mostly involve small or ambiguous objects like plants, clothes, or pictures, which are hard to detect. More robust detection and segmentation models are needed to address this.

\subsection{Ablation Study on Adaptive Exploration Strategy}
\label{subsec:ablation_exploration}

\subsubsection{\textbf{Ablation Study on Exploration Strategy}}
\label{subsubsec:expl strategy}
We conduct an ablation study on HM3Dv2 to evaluate how different exploration strategies affect performance. Four strategies were tested:  
\textit{\textbf{(a)} TSP-Semantic}: selects high-scoring frontiers and solves TSP to plan the order;  
\textit{\textbf{(b)} Greedy-Semantic}: chooses the frontier with the highest semantic score;  
\textit{\textbf{(c)} TSP-Geometry}: solves TSP over all frontiers;  
\textit{\textbf{(d)} Greedy-Geometry}: selects the nearest frontier.
{\textit{\textbf{(e)} Utility-based}: selects the highest utility score (semantic score divided by path length \cite{chen2023traindragontrainingfreeembodied}).}
{We evaluate single-stage strategies (using one method throughout), two-stage strategies (switching between semantic and geometric modes based on thresholds \(\sigma_t = 0.015\) and \(r_t = 1.10\)) and utility-based strategies. Results are shown in Table~\ref{table:ablation_strategy}.}

The results show that two-stage strategies outperform single-stage ones, highlighting the advantage of adaptively switching between semantic and geometric exploration. Among the single-stage strategies, TSP-Semantic and Greedy-Geometry performed best, and their combination achieves the highest overall score: Greedy-Geometry is efficient when semantic cues are weak, while TSP-Semantic guides the agent effectively once strong semantic cues emerge. {Meanwhile, our strategy outperforms utility-based methods by better prioritizing semantically informative regions, even when they are far away, unlike utility-based methods that often undervalue such areas due to longer path costs.}

\subsubsection{\textbf{Ablation Study on Exploration Metrics}}
\label{subsubsec:adaptive metrics}
To explore the balance between semantic-based and geometry-based exploration, we test various combinations of $\sigma_t$ and $r_t$, using SR, SPL, and SoftSPL as evaluation metrics. The results show a hump-shaped curve (see Table~\ref{table:ablation_metrics} and Fig.~\ref{fig:ablation_metrics}). The best performance is achieved with $\sigma_t = 0.015$ and $r_t = 1.10$, supporting the idea that relying too much on either semantic or geometric information reduces efficiency. Instead, an adaptive balance between the two is key to effective exploration.

\begin{table}[t]
    \centering
    \setlength\tabcolsep{3.0pt}
    \resizebox{0.99\columnwidth}{!}{
    \begin{tabular}{lccccccccc}
        \toprule
        \multirow{2}*{\textbf{Method}} & \multicolumn{2}{c}{\textbf{HM3Dv1}} & \multicolumn{2}{c}{\textbf{HM3Dv2}} & \multicolumn{2}{c}{\textbf{MP3D}} \\
        \cmidrule(lr){2-3} \cmidrule(lr){4-5} \cmidrule(lr){6-7}
        & SR$\uparrow$ & SPL$\uparrow$ & SR$\uparrow$ & SPL$\uparrow$ & SR$\uparrow$ & SPL$\uparrow$ \\
        \midrule
        {(a)} w/o Similar Objects & 56.95 & 30.40 & 75.00 & 37.63 & 28.70 & 15.43 \\
        {(b)} w/o Fusion & 44.50 & 25.48 & 55.00 & 29.72 & 27.84 & 14.03 \\
        {(c)} {w/o Fusion (High DCT)} & {55.20} & {31.24} & {68.20} & {34.99} & {37.81} & {17.16} \\
        {(d)} w/o Observation & 50.48 & 29.17 & 70.20 & 37.44 & 35.58 & 16.09 \\
        {(e)} w Max-fusion & 51.43 & 28.97 & 67.90 & 37.04 & 32.11 & 15.32 \\
        \rowcolor{lightgray} \textbf{Ours} & \textbf{59.60} & \textbf{32.95} & \textbf{76.20} & \textbf{38.03} & \textbf{39.23} & \textbf{17.81} \\
        \bottomrule
    \end{tabular}
    }
    \caption{Ablation Study on Semantic Fusion.}
    \label{table:ablation_target-fusion}
    \vspace{-0.5cm}
\end{table}

\subsection{Ablation Study on Target-centric Semantic Fusion}

To assess the impact of the target-centric semantic fusion method, we conduct many ablation experiments. The following control groups were designed: 
\textit{\textbf{(a)} w/o Similar Objects}: Removes the auxiliary detection module for similar objects; \textit{\textbf{(b, c)} w/o Fusion}: Removes the semantic fusion module and directly overwrites cluster confidence using detection results; \textit{\textbf{(b)}} uses a detector confidence threshold (DCT) as in ApexNav, while \textit{\textbf{(c)}} uses a higher DCT as in VLFM. {A lower DCT makes the detector more prone to false positives.} \textit{\textbf{(d)} w/o Observation}: The confidence of the object cluster is not reduced by fusing low-confidence detection results when no objects are detected; \textit{\textbf{(e)} w Max-fusion}: Replaces ApexNav's fusion method with max-confidence fusion.
The results are presented in Table~\ref{table:ablation_target-fusion}.

Both the \textit{\textbf{(b)}} and \textit{\textbf{(c)}} groups underperform compared to Ours, emphasizing the importance of fusion.  {Although \textit{\textbf{(c)}} reduces false detections with a higher DCT, it also misses some correct ones. ApexNav uses a lower DCT and refines more candidates through fusion, leading to better results and showing stronger robustness to false detections.}
Additionally, ApexNav outperforms the max-fusion approach \textit{\textbf{(e)}}, proving its effectiveness in preventing task failures caused by high-confidence mislabeling. The other two experiments emphasize the significance of fusing similar objects and reducing confidence for objects that were previously detected but are currently undetected. Overall, the target-centric semantic fusion method delivers the best performance, demonstrating its effectiveness.

\subsection{Ablation Study on Safe Waypoint Navigation}

To validate the efficiency and safety of ApexNav's navigation strategy, we conduct an ablation study by removing the safe waypoint navigation component and relying solely on the shortest path computed using the A* algorithm for navigation. SR, SPL, and DTG were chosen as evaluation metrics. The experimental results are summarized in Table~\ref{table:ablation_safe-waypoint}.
\begin{table}[t]
    \centering
    \setlength\tabcolsep{7pt}
    \resizebox{0.8\columnwidth}{!}{
    \begin{tabular}{lccc}
         \toprule
         \textbf{Navigation Method} & SR$\uparrow$ & SPL$\uparrow$ & DTG$\downarrow$ \\
         \midrule
         The Shortest Path & 66.20 & 33.85 & 1.72 \\
         \rowcolor{lightgray} \textbf{Safe Navigation (Ours)} & \textbf{76.20} & \textbf{38.03} & \textbf{1.31} \\
         \bottomrule
    \end{tabular}
    }
    \caption{\textbf{Ablation Study on Safe Waypoint Navigation.} This table compares the result with "The Shortest Path" on the HM3Dv2 dataset.}
    \label{table:ablation_safe-waypoint}
    \vspace{-0.5cm}
\end{table}

The results reveal that relying solely on the shortest path for navigation leads to a notable decline in SR and SPL, accompanied by a significant increase in DTG.
This indicates that the agent's traversal paths deteriorate, often resulting in collisions that disrupt stable progress toward the target. These findings highlight the effectiveness of the safe waypoint navigation method in enhancing both efficiency and safety.

\subsection{Real-world Deployment}

We conduct real-world experiments using the AgileX LIMO robot. A laptop equipped with an RTX 4060 Laptop GPU and an AMD R9 7945HX CPU served as the ROS master and communicated with the robot via a ROS-based master-slave architecture. We use MINCO \cite{wang2022geometrically} to generate collision-free spatiotemporal trajectories for waypoint navigation. {ApexNav is evaluated in dormitory and laboratory environments across six trials, targeting toilet, couch, and chair in the former, and refrigerator, plant, and TV in the latter.} 
{Representative results and quantitative statistics are shown in Fig. \ref{fig:realworld} and Table \ref{tab:real_world_data}.} 
{Module runtimes are summarized in Table~\ref{tab:run_time_all}. Most modules run within the desired interval, even on a consumer-grade laptop. Although detection slightly exceed this range, it has little effect on overall task performance. This further highlights the efficiency and deployability in practical scenarios.}

\begin{table}[t]
    \centering
    \setlength\tabcolsep{2.0pt}
    \renewcommand{\arraystretch}{1.1}
    \begin{subtable}{\linewidth}
        \centering
        \resizebox{0.95\columnwidth}{!}{
        \begin{tabular}{cccc}
            \hline
            \multicolumn{1}{c}{\multirow{2}{*}{\textbf{Modules}}} & \multicolumn{2}{c}{\textbf{GPU Runtime (ms)}} & \multicolumn{1}{c}{\textbf{Desired}}\\ \cline{2-3}
            \multicolumn{1}{c}{} & RTX 4060 Laptop & RTX 4090 & \textbf{Interval (ms)}\\ \hline
            Detection+Segmentation & $[190,320]$ & $[65,125]$  & 250\\ 
            BLIP2 & $[160,190]$ & $[50,55]$ & 200\\ \hline
        \end{tabular}
        }
        \caption{{Runtime of modules on different GPU devices.}}
        \label{tab:run_time_gpu}
    \end{subtable}
    \begin{subtable}{\linewidth}
        \centering
        \resizebox{0.99\columnwidth}{!}{
        \begin{tabular}{cccc}
            \hline
            \multicolumn{1}{c}{\multirow{2}{*}{\textbf{Modules}}} & \multicolumn{2}{c}{\textbf{CPU Runtime (ms)}} & \multicolumn{1}{c}{\textbf{Desired}}\\ \cline{2-3}
            \multicolumn{1}{c}{} & AMD Ryzen9 7945hx & Intel i9-13900K & \textbf{Interval (ms)} \\ \hline
            Semantic Object Map & $[15,20]$ & $[5,15]$ & 250\\ 
            Semantic Score Map & $[1,2]$ & $[0.5,1]$ & 200 \\
            Grid Map & $[30,50]$ & $[20,35]$ & 100\\
            ESDF Map & $[1,3]$ & $[1,2]$ & 100 \\
            Exploration Planning & $[50, 200]$ & $[40, 150]$ & 500 \\
            \hline
        \end{tabular}
        }
        \caption{{Runtime of modules on different CPU devices.}}
        \label{tab:run_time_cpu}
    \end{subtable}
    \caption{{\textbf{Time consumption of modules on different devices.} The \textit{Desired Interval} indicates the runtime range required for stable real-time performance.}}
    \label{tab:run_time_all}
    \vspace{-0.10cm}
\end{table}

\begin{table}[t]
    \centering
    \setlength\tabcolsep{4.0pt}
    \resizebox{0.35\textwidth}{!}{
    \begin{tabular}{ccccc}
        \hline
        \textbf{Trials} & \textbf{SR (\%)} & \textbf{SPL} & \textbf{TD (m)} & \textbf{TT (s)} \\
        \hline
        6 & 100 & 0.723 & 15.85 & 56 \\
        \hline
    \end{tabular}
    }
    \caption{{\textbf{Quantitative Statistics in Real-world Experiments} Results are averaged over all trials. TD: Traveled Distance; TT: Traveled Time.}}
    \label{tab:real_world_data}
    \vspace{-0.5cm}
\end{table}


\section{Conclusion} 
\label{sec:conclusion}

In this paper, we propose ApexNav, an efficient and reliable ZSON framework. ApexNav features an adaptive exploration strategy that dynamically combines semantic- and geometry-based approaches, enabling the agent to effectively balance between them based on the reliability of semantic cues. In addition, we propose a target-centric semantic fusion method that preserves long-term memory of the target object and its semantically similar counterparts, effectively mitigating task failures caused by misidentification and significantly improving success rates.
ApexNav achieves SOTA performance on three benchmark datasets and is further validated through extensive real-world experiments.

While ApexNav performs well, it has several limitations worth exploring. First, it assumes the target is visible, but objects may be hidden (e.g., in drawers), requiring interaction or manipulation.
In real-world tests, we also observed two issues. One is that the system assumes all areas in the sensor’s view are fully explored, which may not hold for small objects that require finer-grained exploration. Another is that when the target has weak semantic relevance to its surroundings, semantic guidance becomes ineffective, leading to inefficient geometric exploration. Using richer language priors, as in Vision-Language Navigation, may help overcome this.


{\footnotesize
\bibliographystyle{plainnat}
\bibliography{references}
}

\newpage
\clearpage
\appendix

\subsection{LLM Prompt}
\label{subsec:LLM Prompt}

In Sec. \ref{subsubsec:semantic_score_map}, we leverage LLM inference to determine the most likely room where the target object appears. In Sec. \ref{subsubsec:Object Reasoning with LLMs}, we utilize LLMs to infer the fusion confidence threshold for the object goal and generate a list of similar objects. The complete LLM prompt is as follows:

\textbf{Step 1 (System set of the LLM):} In this step, we set the system of LLM and its task.

\begin{graybox}
    \hspace{1em}You are an intelligent agent that helps the object detector identify detected objects. Based on the labels I provide, you'll give the labels that may cause misdetection, potential room objects, and the confidence threshold.

    At each step, I will give you a label.You should consider the following:

    (1) For misdetection, provide objects that may be misdetected by detectors such as YOLO and GroundingDINO, considering similarities in shape, color, size, etc. The labels should be in COCO classes, with at least 3 and at most 5 labels.

    (2) For the confidence score, the confidence score should be in [0.25, 0.65], as the smallest is an apple in a home (0.25) and the largest is a bed in a home (0.65).

    (3) For the room, guess where the object is most likely to appear, with a significantly higher probability than in other rooms. If the probabilities are similar across rooms, return "everywhere."
\end{graybox}

\textbf{Step 2 (Few-shot prompts):} In this step, we give LLM two examples for fine-tuning. USET1 and ASSISTENT1, and USET2 and ASSISTENT2 are two sets of examples given to LLM, training it how to think and respond.

\begin{graybox}
    \hspace{1em}\textbf{USER1} = cake

    \textbf{ASSISTANT1} = Thought: In the COCO classes, confidence threshold for cake: A cake is typically round or square and can have decorative icing or layers. It is quiet smaller than a bed (0.65), but bigger than an apple (0.25), so the confidence threshold should be low, around 0.30, which is in [0.25, 0.65].

    Misdetection possibilities:

    donut: A donut, especially a large one or a stack of donuts, can resemble a cake because of its circular shape and icing-like decorations.

    pizza: A whole pizza, particularly one with thick crust or layered toppings, might resemble a cake in certain contexts.

    sandwich: A stacked sandwich, especially one with visible layers, could look similar to a layered cake. 

    pie: A pie can be misdetected as a cake due to its circular shape and pastry-like crust, which is similar to a cake's appearance.

    Room where the label may appear: A cake can be in a kitchen or a living room.The number of rooms is more than 1, so a cake can be "everywhere".

    Answer: [donut, pizza, sandwich, pie, 0.30, everywhere]

    \textbf{USER2} = "dining table"

    \textbf{ASSISTANT2} = Thought: In the COCO classes, confidence threshold for cabinet: A cabinet is typically rectangular, with flat surfaces, often in the form of a box or with shelves inside. It’s smaller than larger furniture pieces like beds or dining tables (0.60) but still sizable compared to smaller household objects. The confidence threshold for a cabinet should be around 0.50, which is in [0.25, 0.65].
    
    Misdetection possibilities:
    
    bookshelf: A bookshelf can be similar in shape and size to a cabinet, with a flat surface and storage space, which may lead to confusion.
    
    dresser: A dresser, especially one with drawers, could resemble a cabinet because of the similar rectangular shape and flat surfaces.
    
    closet: A closet, particularly a small one, can resemble a cabinet in terms of its rectangular form and purpose for storage.
    
    Room where the label may appear: A cabinet is most likely to appear in a kitchen, bathroom, or hallway for storage purposes.
     
    It could also be found in a living room, depending on the layout of the space.
    
    Answer: [bookshelf, dresser, closet, 0.50, kitchen]
\end{graybox}

\textbf{Example:} In the following, We show a successful communitaction. INPUT is the input label of the episode, and RESPONSE is the output of the LLM.

\begin{graybox}
    \hspace{1em}\textbf{INPUT} = toilet

    \textbf{RESPONSE} = Thought: In the COCO classes, confidence threshold for toilet: A toilet has a distinct shape, but it is relatively smaller than larger furniture like beds or couches. It's smaller than a dining table (0.60) but still larger than small items like an apple (0.25). The confidence threshold for a toilet should be around 0.45, which is in [0.25, 0.65].
    
    Misdetection possibilities:
    
    chair: A low chair, especially one with a rounded backrest, could be misdetected as a toilet from certain angles.
    
    sink: A sink, particularly those with a base and a bowl-like shape, might resemble a toilet if only partially visible.
    
    potted plant: A large potted plant on a stand might look similar to a toilet due to its size and cylindrical form, though this is less common.
    
    Room where the label may appear: A toilet is most likely to appear in a bathroom. It is almost never found outside of this context, as its usage is confined to this room.
    
    Answer: [chair, sink, potted plant, 0.45, bathroom]
\end{graybox}

In this paper, we made some adjustments to the labels to ensure that the detector's detection results better align with the requirements of the dataset. 

For HM3D-v0.1 and HM3D-v0.2, we set the confidence threshold range to [0.25, 0.65]. The inference results of the LLM we used are as follows:

\begin{bluebox}
    \noindent \textbf{couch}: ['chair', 'bed', 'bench', 'dining table', 0.6, 'living room'] \# sofa
    
    \noindent \textbf{tv}: ['laptop', 'picture frame', 'window', 0.45, 'living room'] \# tv monitor
    
    \noindent \textbf{chair}: ['couch', 'toilet', 'potted plant', 0.5, 'everywhere']
    
    \noindent \textbf{toilet}: ['chair', 'bench', 'potted plant', 'sink', 0.45, 'bathroom']
    
    \noindent \textbf{bed}: ['couch', 'dining table', 'bench', 0.65, 'bedroom']
    
    \noindent \textbf{potted plant}: ['lamp', 'teddy bear', 'toilet', 0.4, 'everywhere'] \# plant
\end{bluebox}

For MP3D, some labels in the dataset are uncommon, and certain scene and object shapes in MP3D are incomplete, making it difficult for the detector to identify the targets. Therefore, we set the confidence threshold range to [0.25, 0.35] for the objects that are harder to detect. The inference results of the LLM we used are as follows:

\begin{bluebox}
    \noindent \textbf{couch}: ['bed', 'bench', 'chair', 'dining table', 0.34, 'living room'] \# sofa

    \noindent \textbf{tv}: ['laptop', 'monitor', 'picture frame', 'window', 0.27, 'living room'] \# tv monitor
    
    \noindent \textbf{chair}: ['couch', 'toilet', 'stool', 'potted plant', 0.30, 'everywhere']
    
    \noindent \textbf{toilet}: ['chair', 'bench', 'potted plant', 'sink', 0.27, 'bathroom']
    
    \noindent \textbf{bed}: ['couch', 'dining table', 'bench', 0.35, 'bedroom']
    
    \noindent \textbf{potted plant}: ['lamp', 'teddy bear', 'toilet', 0.25, 'everywhere'] \# plant
    
    \noindent \textbf{cabinet}: ['refrigerator', 'door', 0.32, 'everywhere']
    
    \noindent \textbf{table}: ['bed', 'couch', 'bench', 'counter', 0.30, 'everywhere']
    
    \noindent \textbf{pillow}: ['suitcase', 'backpack', 'handbag', 'teddy bear', 0.25, 'everywhere'] \# cushion
    
    \noindent \textbf{counter}: ['dining table', 'bed', 'couch', 'bench', 0.30, 'living room']
    
    \noindent \textbf{sink}: ['toilet', 'bathtub', 'washing machine', 'refrigerator', 0.27, 'everywhere']
    
    \noindent \textbf{framed photograph}: ['tv', 'laptop', 'book', 'clock', 0.27, 'everywhere'] \# picture
    
    \noindent \textbf{fireplace}: ['oven', 'refrigerator', 'tv', 0.30, 'living room']
    
    \noindent \textbf{towel}: ['blanket', 'curtain', 'rug', 0.25, 'bathroom']
    
    \noindent \textbf{seating}: ['couch', 'bed', 'potted plant', 'dining table', 0.30, 'everywhere']
    
    \noindent \textbf{nightstand}: ['bench', 'chair', 'stool', 0.32, 'bedroom'] \# chest of drawers
    
    \noindent \textbf{shower}: ['toilet', 'sink', 'refrigerator', 0.25, 'bathroom']
    
    \noindent \textbf{bathtub}: ['bed', 'couch', 'dining table', 'potted plant', 0.30, 'bathroom']
    
    \noindent \textbf{clothes}: ['towel', 'blanket', 'curtain', 0.27, 'everywhere']
    
    \noindent \textbf{stool}: ['chair', 'potted plant', 'toilet', 0.28, 'everywhere']
    
    \noindent \textbf{gym equipment}: ['bench', 'chair', 'sports ball', 'dumbbell', 0.28, 'everywhere']
\end{bluebox}
The bolded labels are used in the experiment. Labels with " \# label" are the original dataset labels, now replaced by the bolded ones. Other labels are the original dataset labels that have not changed.

\subsection{Details of Failure Cause Analysis}
\label{subsec:details of failure cause analysis}

\begin{figure}[h]
    \begin{center}
      \includegraphics[width=0.74\columnwidth]{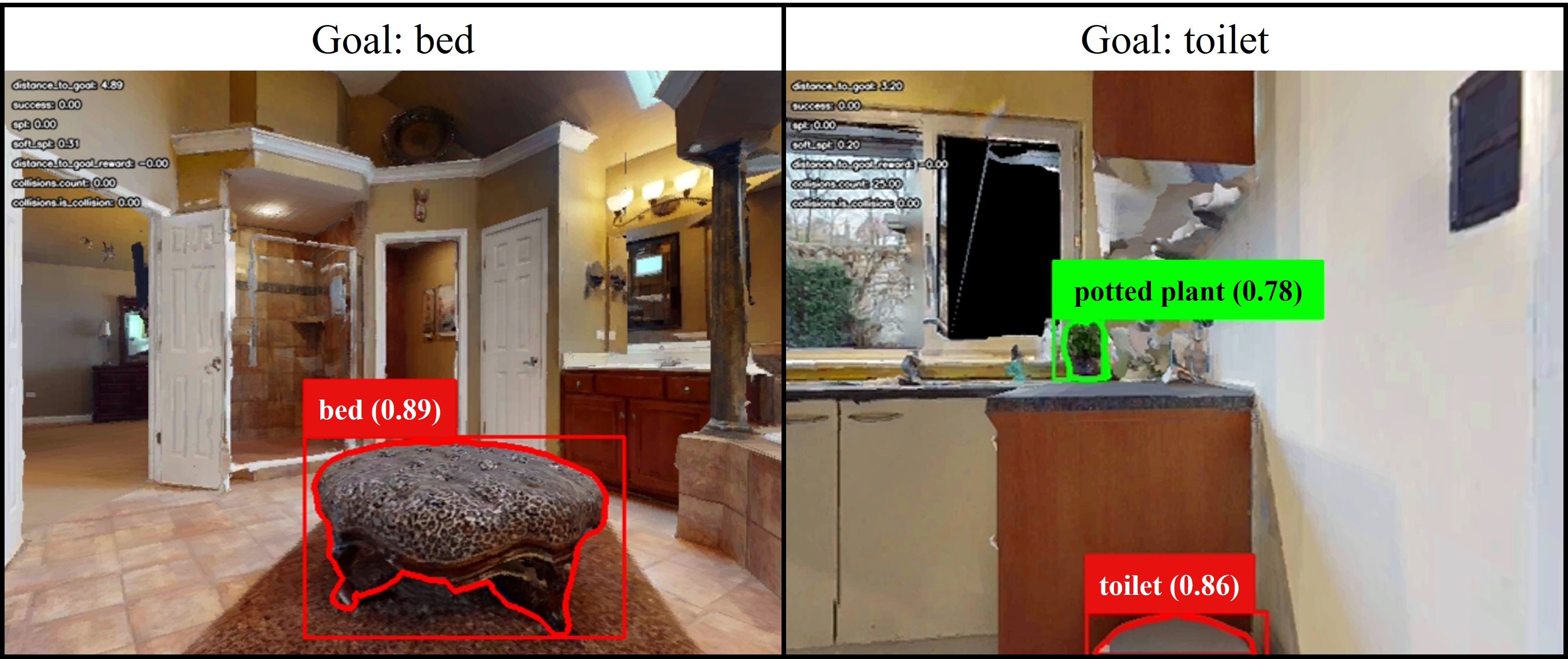} 
    \end{center}
   \vspace{-0.3cm}
   \caption{\label{fig:detection errors} Detection Errors. }
   \vspace{-0.65cm}
\end{figure}

As shown in Fig.~\ref{fig:detection errors}, even with the addition of the target-centric semantic fusion module, the performance limitations of the detectors still lead to the misdetection of certain incorrect objects as task targets, resulting in task failure.

\begin{figure}[h]
    \begin{center}
      \includegraphics[width=0.74\columnwidth]{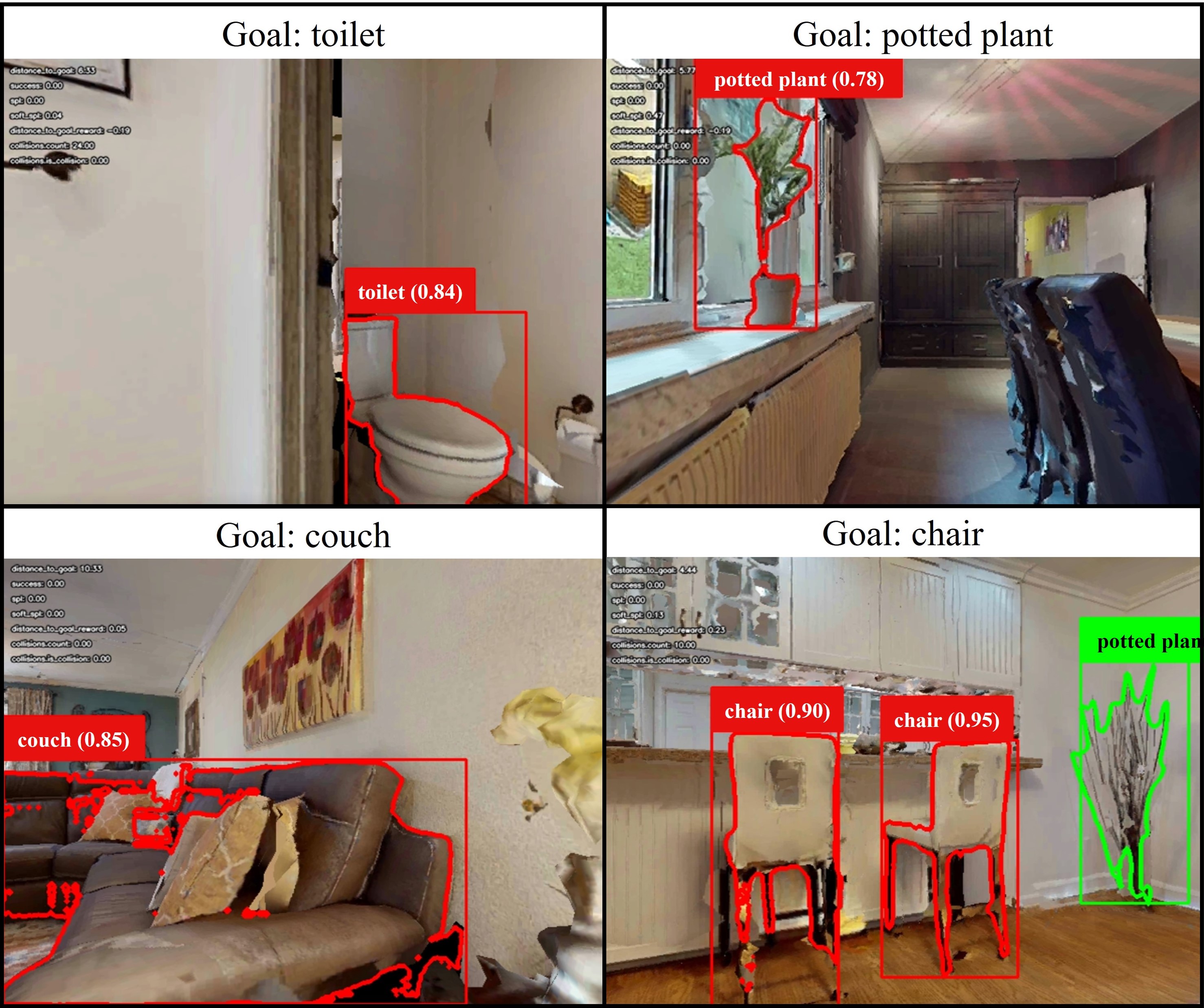} 
    \end{center}
   \vspace{-0.3cm}
   \caption{\label{fig:no labels} Objects with Incorrect Labels. }
   \vspace{-0.5cm}
\end{figure}

A more significant issue is that there are still some objects with incorrect labels in the datasets. Fig.~\ref{fig:no labels} highlights several misclassified objects in HM3D-v0.1. Although these objects are consistent with human recognition, they were not annotated as target objects in the HM3D-v0.1 task annotations. Many of these labeling issues have been addressed in the HM3D-v0.2 dataset.

\subsection{Details of Ablation Experiments}
\label{subsec:details of ablation experiments}

To further demonstrate the superiority of target-centric semantic fusion, this study will conduct a detailed analysis of the ablation group where the success rate significantly decreases.

\subsubsection{ApexNav w/o Fusion on HM3D-v0.2}
\label{subsubsec:ApexNav w/o Fusion on HM3D-v0.2}

\begin{figure}[t]
    \centering
        \begin{subfigure}[b]{\columnwidth}
            \centering
            \includegraphics[width=0.9\textwidth]{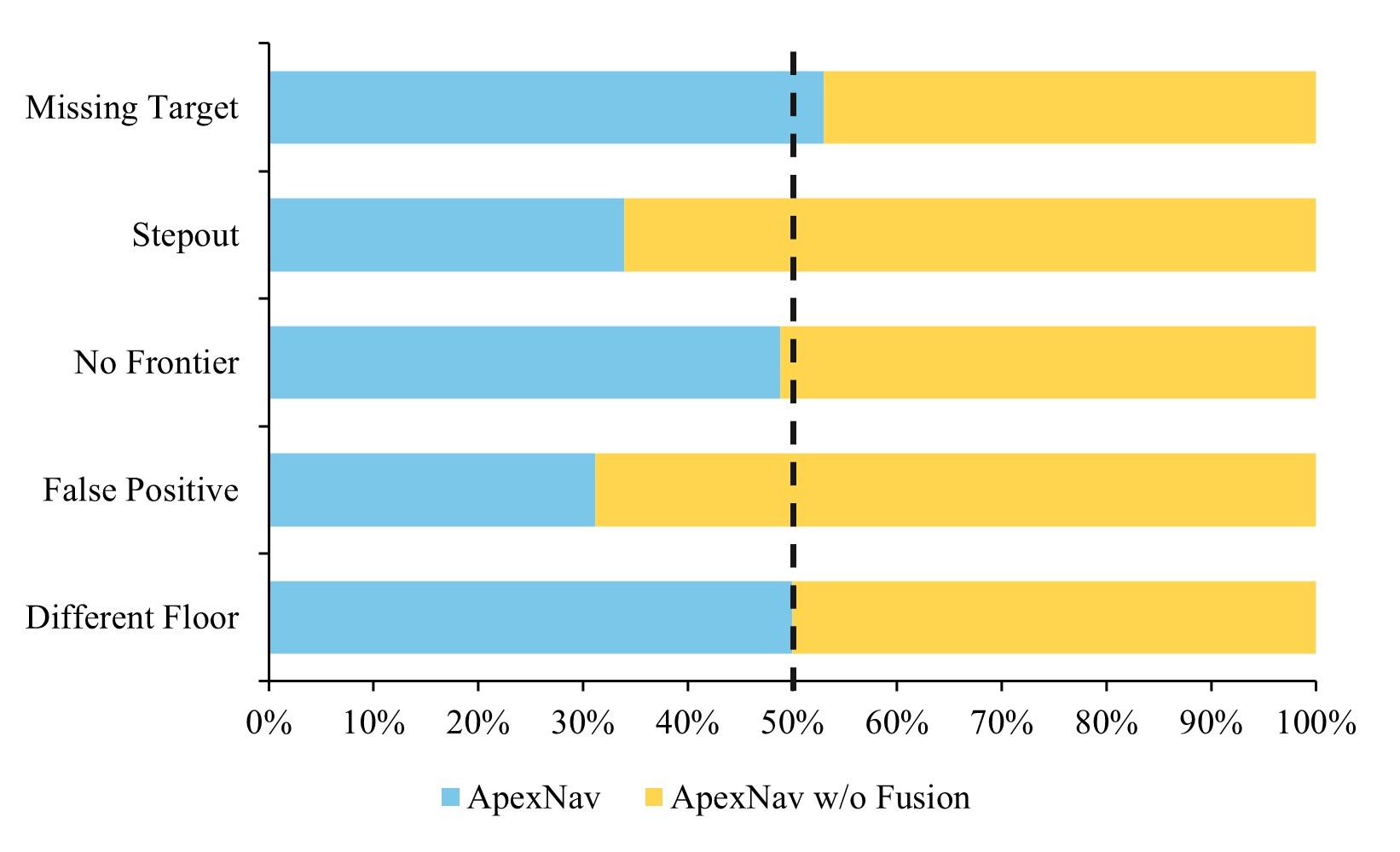}
            \vspace{-0.3cm}
            \caption{Comparison of Failure Causes.}
            \label{fig:comparison of failure counts of w/o fusion}
        \end{subfigure}

        \begin{subfigure}[b]{\columnwidth}
            \centering
            \includegraphics[width=0.9\textwidth]{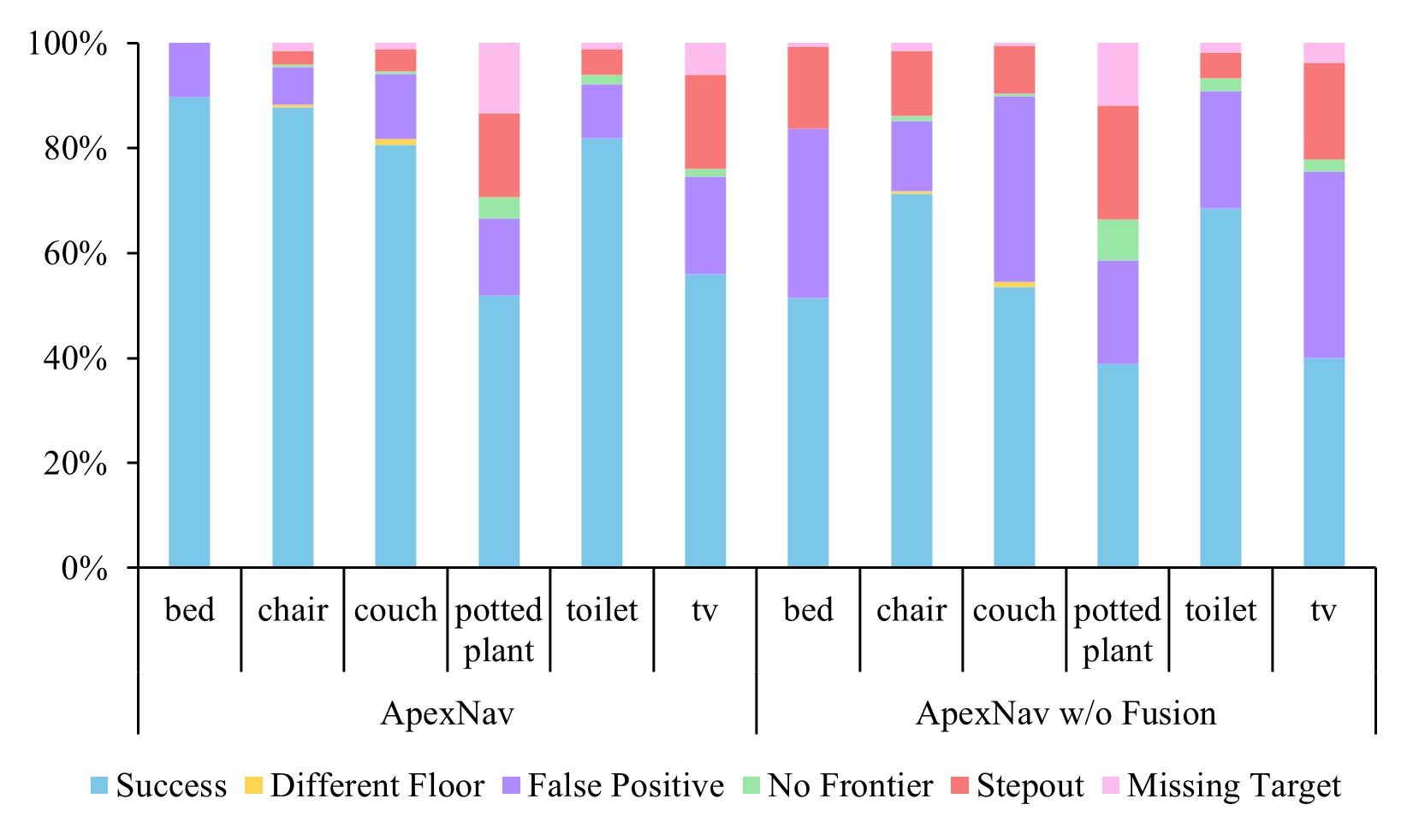}
            \vspace{-0.3cm}
            \caption{Comparison of Results Across Categories.}
            \label{fig:results counts in hm3dv2}
        \end{subfigure}
    \caption{Comparisons with ApexNav w/o Fusion Group on HM3D-v0.2 Dataset.}
    \label{fig:comparison of w/o fusion}
    \vspace{-0.3cm}
\end{figure}

Compared to ApexNav, the ApexNav w/o fusion group shows a very significant decrease in SR across all three datasets. We choose the HM3D-v0.2 dataset, which shows the greatest decrease in SR, for analysis.
According to Fig.~\ref{fig:comparison of w/o fusion}, we found that the ApexNav w/o fusion group exhibited more "False Positive" and "Stepout" cases, leading to the decrease in SR.

\begin{figure}[h]
    \centering
        \begin{subfigure}[b]{\columnwidth}
            \centering
            \includegraphics[width=0.9\columnwidth]{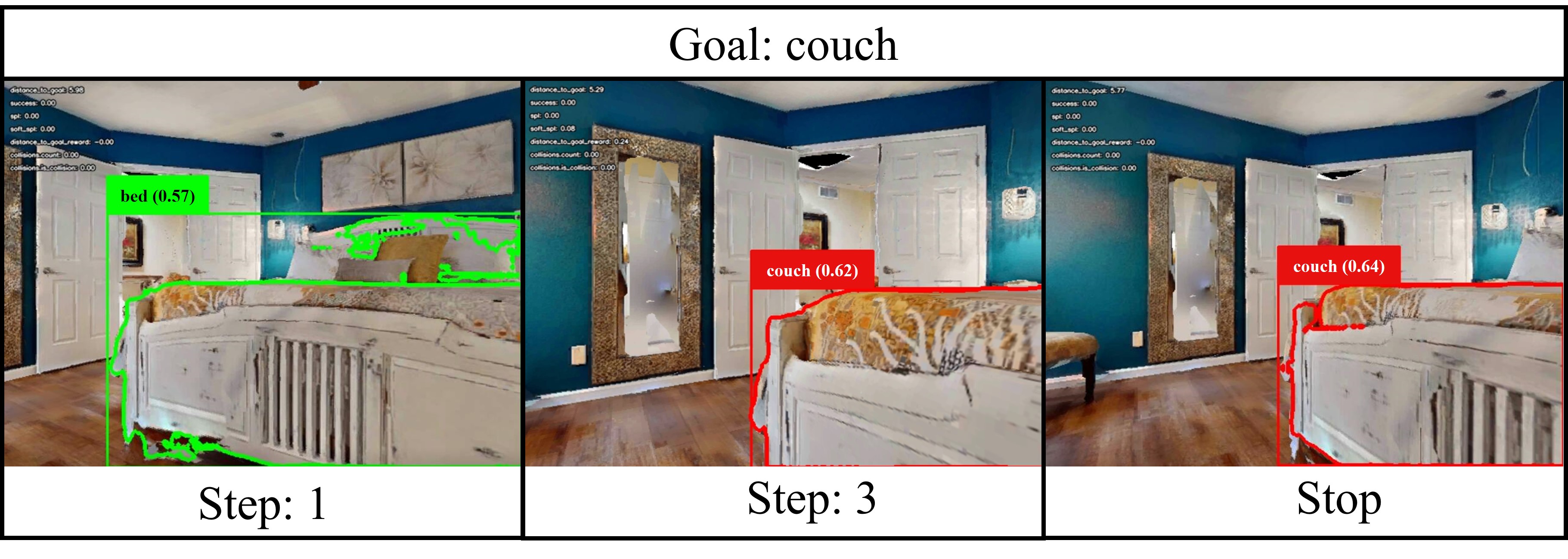} 
            \vspace{-0.3cm}
            \caption{\label{fig:no fusion false positive} False Positive  }
        \end{subfigure}

        \vspace{0.3cm}

        \begin{subfigure}[b]{\columnwidth}
            \centering
            \includegraphics[width=0.9\columnwidth]{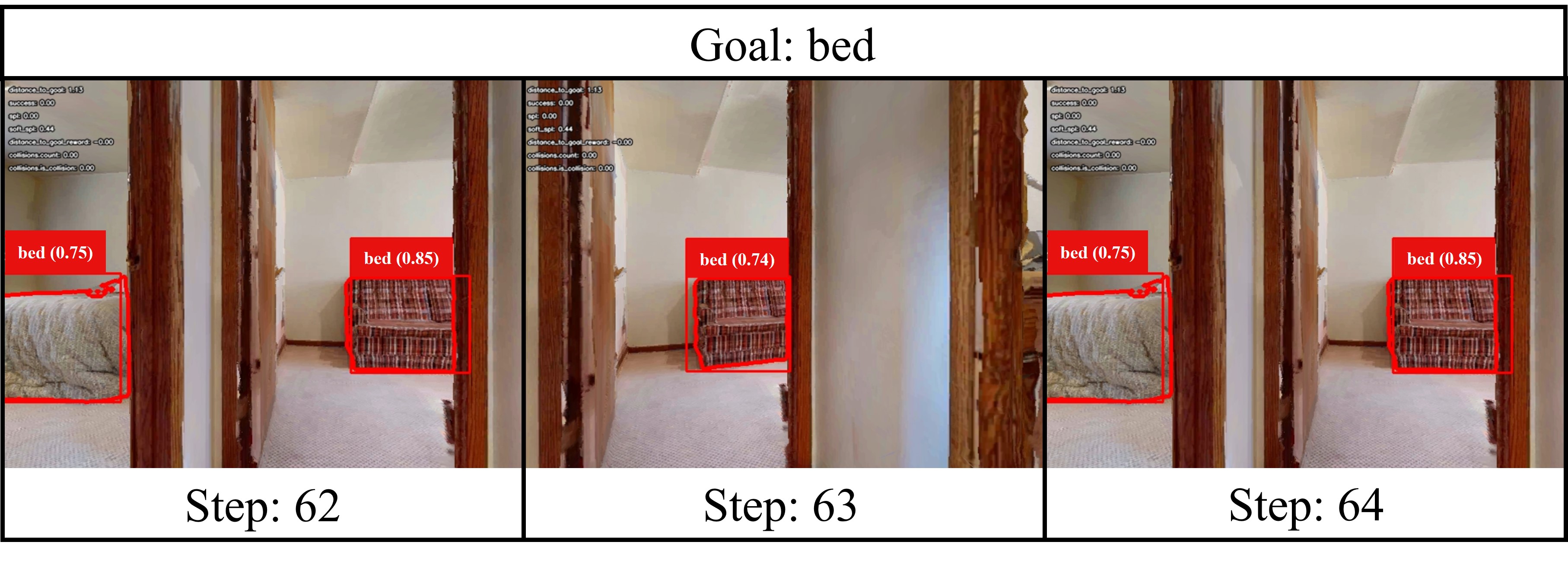} 
            \vspace{-0.3cm}
            \caption{\label{fig:no fusion stepout} Stepout }
        \end{subfigure}
    \caption{Two Examples of "False Positive" and "Stepout" in ApexNav w/o Fusion Group.}
    \label{fig:comparison of w/o Fusion}
    \vspace{-0.3cm}
\end{figure}

For "False Positive" failures, as shown in Fig.~\ref{fig:no fusion false positive}, the agent relies solely on the detector's current results. During exploration, if the agent sees only part of a non-target object, it may misclassify it as the target, ignoring prior detections. If close enough, the agent may mistakenly issue a "STOP" signal, causing a "False Positive" failure.

For "Stepout" failures, as shown in Fig.~\ref{fig:no fusion stepout}, the agent frequently turns left-right due to fluctuating highest-confidence objects between frames. Since ApexNav w/o fusion selects only the highest-confidence object, this inconsistency prevents the agent from choosing a stable navigation target, leading to excessive steps and a "Stepout" failure.

\subsubsection{ApexNav w/o Similar Objects on MP3D}
\label{subsubsec:ApexNav w/o Similar Objects on MP3D}

\begin{figure}[t]
    \centering
        \begin{subfigure}[b]{\columnwidth}
            \includegraphics[width=0.85\textwidth]{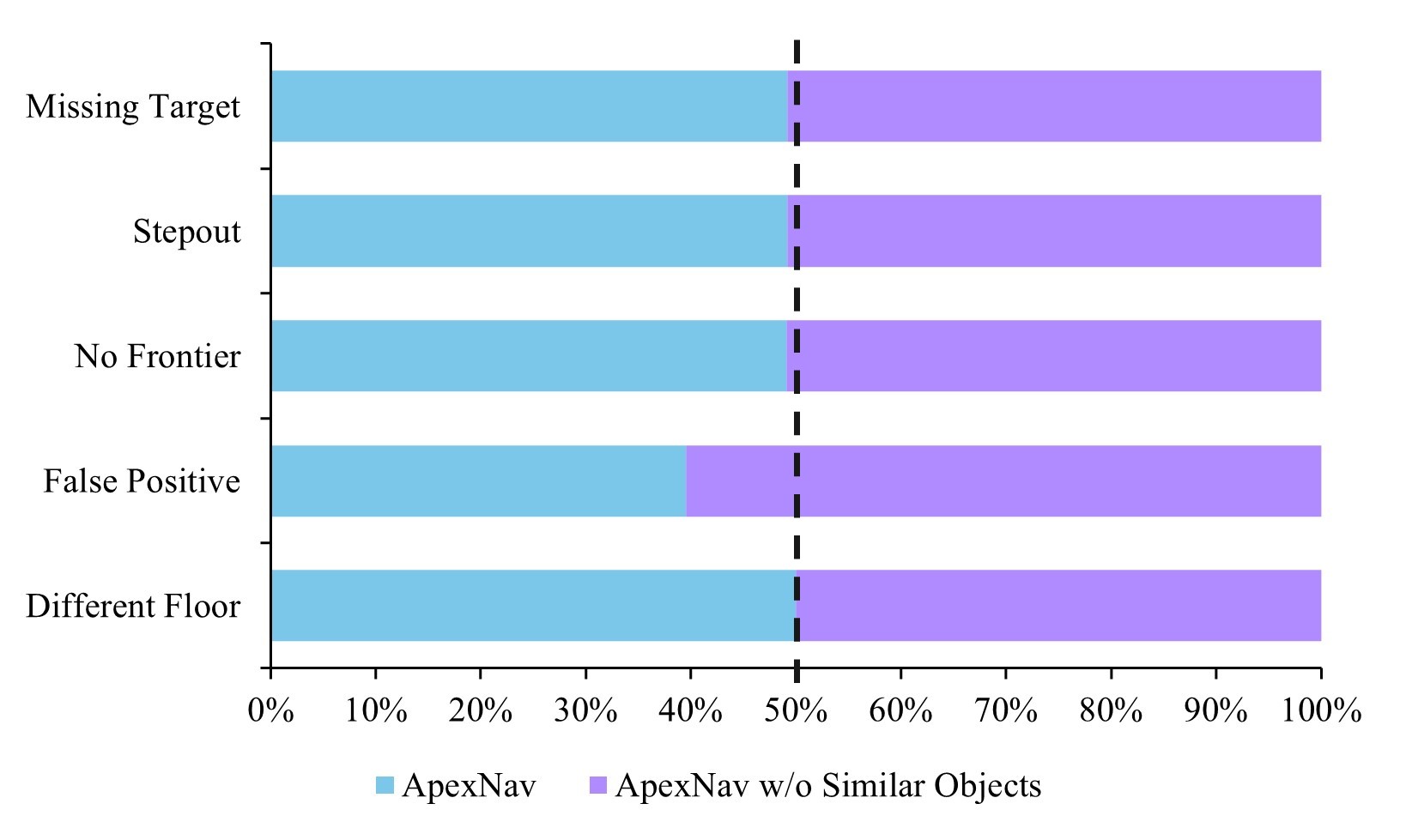}
            \caption{Comparison of Failure Causes}
            \label{fig:comparison of failure counts of w/o similar objects}
        \end{subfigure}

        \vspace{0.1cm}

        \begin{subfigure}[b]{\columnwidth}
            \includegraphics[width=0.85\textwidth]{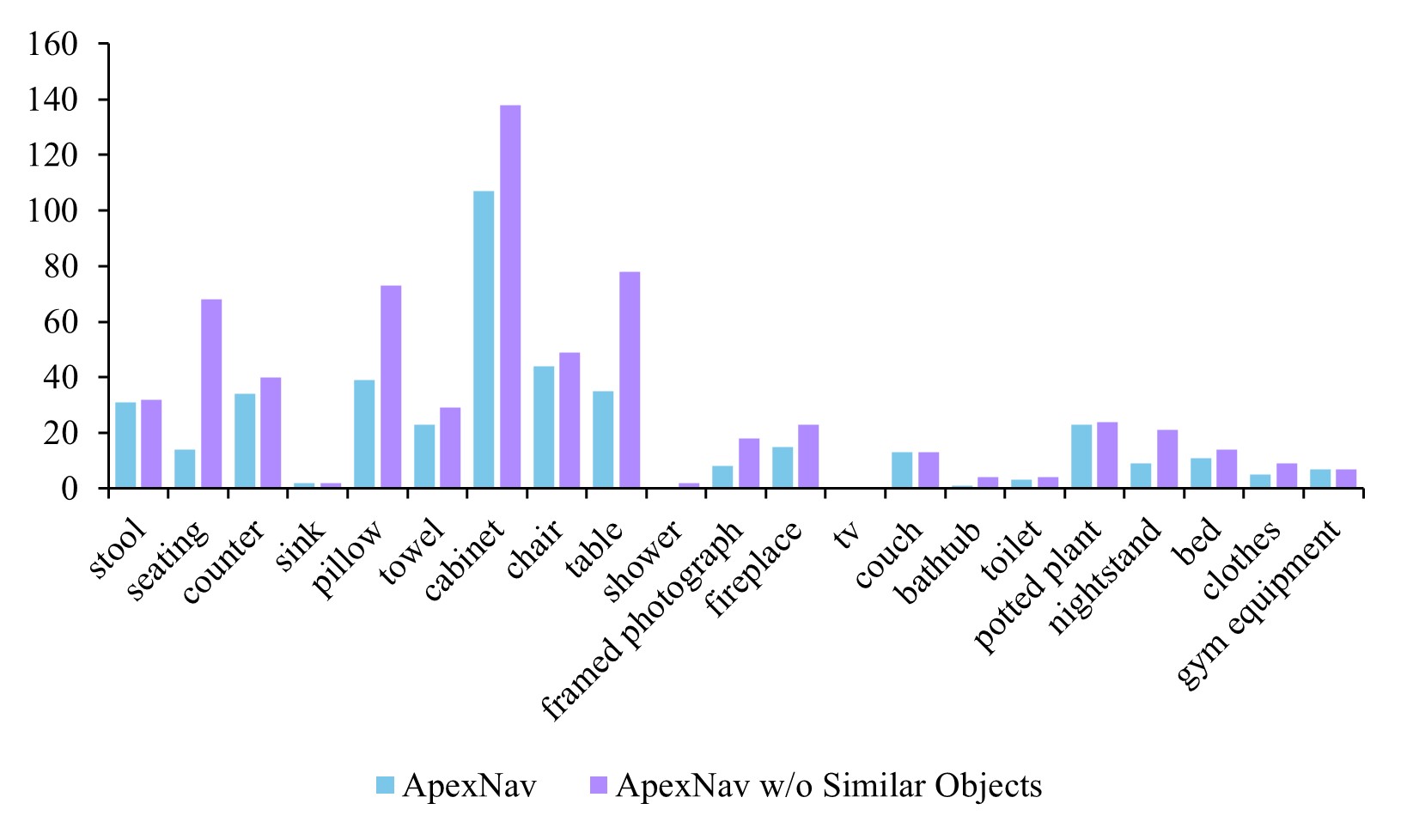}
            \vspace{-0.1cm}
            \caption{Comparison of False Positive for Labels}
            \label{fig:numbers of false positive in MP3D}
        \end{subfigure}
    \caption{Comparison with ApexNav w/o Similar Objects on MP3D dataset.}
    \label{fig:comparison of w/o similar objects}
    \vspace{-0.4cm}
\end{figure}

\begin{figure}[h]
    \begin{center}
      \includegraphics[width=0.9\columnwidth]{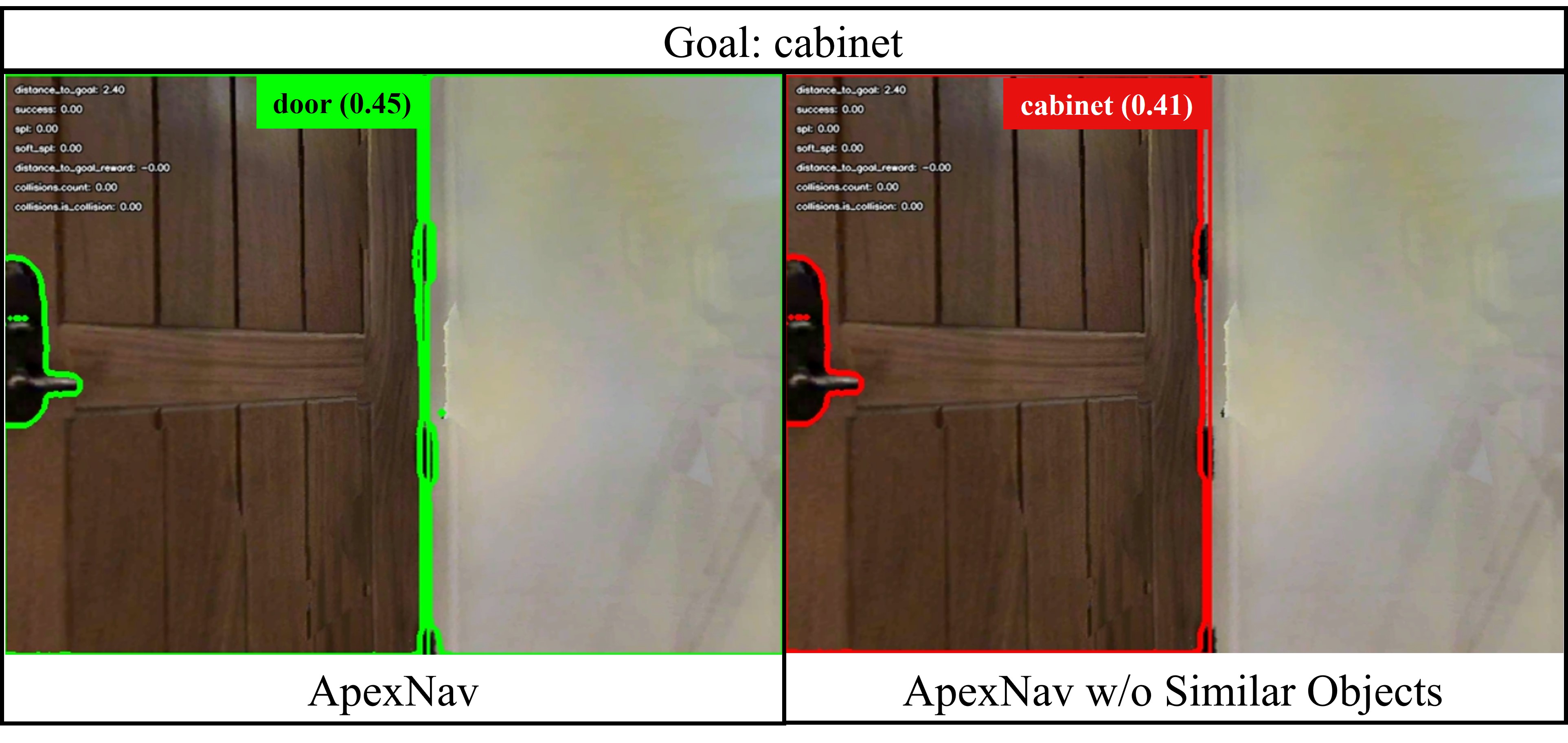} 
    \end{center}
   \vspace{-0.3cm}
   \caption{\label{fig:cabinet detection errors} Comparison Example: ApexNav vs. ApexNav w/o Similar Objects. }
   \vspace{-0.3cm}
\end{figure}

Compared to the results using auxiliary detection based on similar objects, the ApexNav w/o similar objects group shows a significant decline in SR on the MP3D dataset. As shown in Fig.~\ref{fig:comparison of w/o similar objects}, this drop is primarily due to the presence of targets with ambiguous features (e.g., cabinets in Fig.~\ref{fig:cabinet detection errors}), which increases the likelihood of misidentifying similar objects as the target, leading to "False Positives." The method leveraging auxiliary detection based on a list of similar objects helps correct these misclassifications, effectively reducing "False Positives." This highlights the effectiveness of incorporating similar object detection in ApexNav's fusion method.

\end{document}